%% file: BHTS.tex
\documentclass[11pt,a4paper]{article}

\usepackage{graphicx}
\usepackage{epsfig} % less modern
\usepackage[english]{babel}
\usepackage{url}
\usepackage{bm}
\usepackage{color}
\usepackage{lineno}
\usepackage{authblk}
\usepackage[square,sort,comma,numbers]{natbib}

\providecommand{\keywords}[1]{\textbf{\textit{Key Words:}} #1}

\begin{document}
\title{Bayesian Multi Plate High Throughput Screening of Compounds}
\author[1]{Ivo D. Shterev\thanks{Corresponding Author: Ivo D. Shterev, PhD; email: {\tt i.shterev@duke.edu}}}
\author[2]{David B. Dunson}
\author[3]{Cliburn Chan}
\author[1]{Gregory D. Sempowski}
\affil[1]{Duke Human Vaccine Institute, Duke University}
\affil[2]{Department of Statistical Science, Duke University}
\affil[3]{Department of Biostatistics and Bioinformatics, Duke University}
\maketitle

%\linenumbers
\abstract{High throughput screening of compounds (chemicals) is an essential part of drug discovery \citep{malo}, involving thousands to millions of compounds, with the purpose of identifying candidate hits. Most statistical tools, including the industry standard B-score method, work on individual compound plates and do not exploit cross-plate correlation or statistical strength among plates. We present a new statistical framework for high throughput screening of compounds based on Bayesian nonparametric modeling. The proposed approach is able to identify candidate hits from multiple plates simultaneously, sharing statistical strength among plates and providing more robust estimates of compound activity. It can flexibly accommodate arbitrary distributions of compound activities and  is applicable to any plate geometry. The algorithm provides a principled statistical approach for hit identification and false discovery rate control. Experiments demonstrate significant improvements in hit identification sensitivity and specificity over the B-score method, which is highly sensitive to threshold choice. The framework is implemented as an efficient {\tt R} extension package {\tt BHTSpack} and is suitable for large scale data sets.}


\section{Introduction}
\begin{figure}[h]
\centering\scalebox{0.5}{
\includegraphics{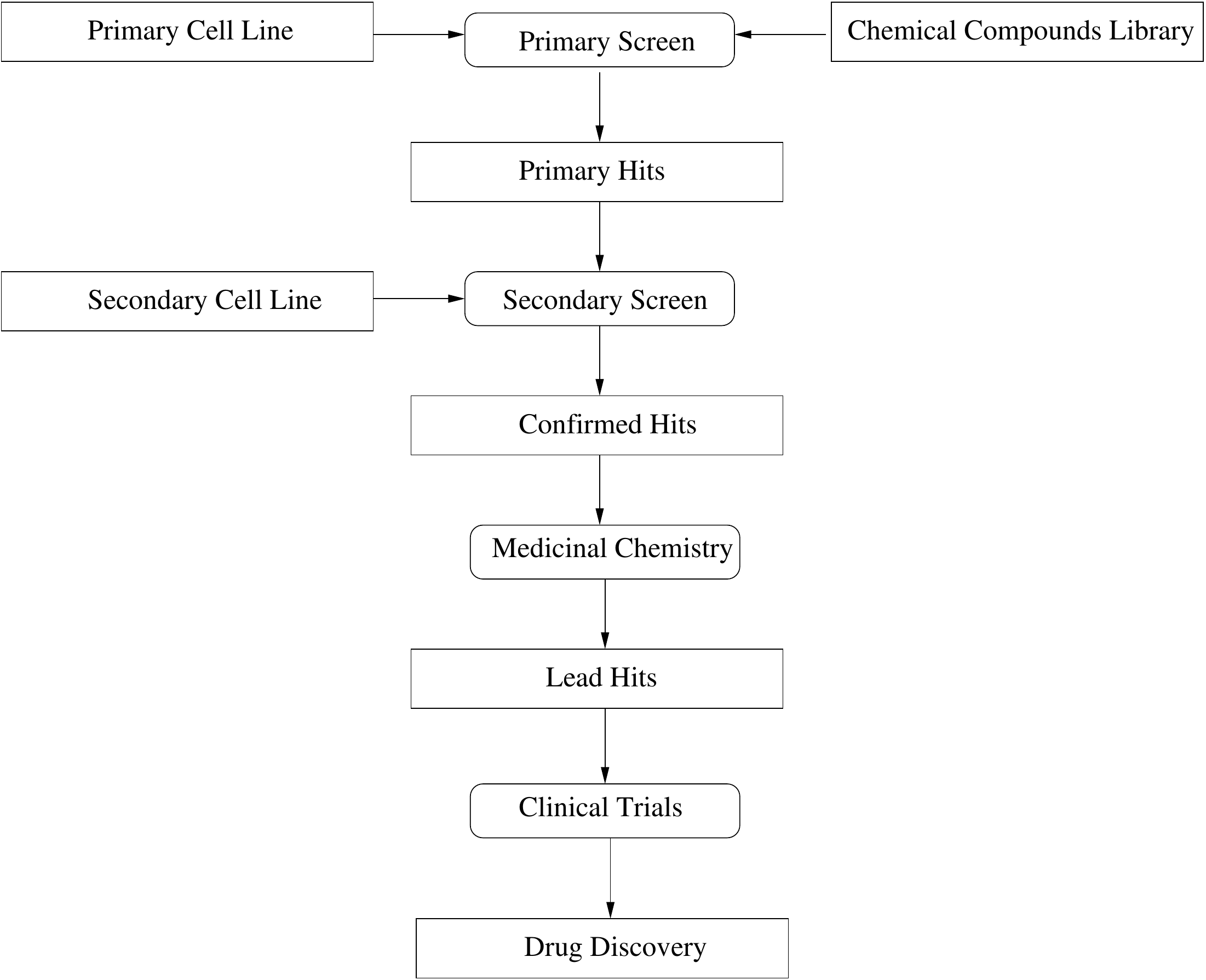}}
\caption{Block diagram of an HTS process.}
\label{HTSprocess}
\end{figure}

High-throughput screening (HTS) of compounds is a critical step in drug discovery \citep{malo}. This typically involves the screening of thousands to millions of candidate compounds (chemicals). The objective is to accurately identify which compounds are candidate active compounds (hits). Those compounds will then undergo a secondary screen. A flow chart of a typical HTS process is shown in Fig. \ref{HTSprocess}. The first step in the process, called primary screening, is a comprehensive scan of tens of thousands of compounds with the objective of identifying primary hits. Most often the compounds are run in singlets, although a more desired experimental design will involve replicates. Computational and statistical tools involved in the primary screening step need to be accurate and also efficient, due to the large number of compounds to be screened.

Two types of error can occur in the primary screening process, namely false positive (FP) and false negative (FN) errors. While technological improvements and advances in experimental design and accuracy can help mitigate these two types of error, they by themselves are not able to sufficiently improve the quality of the HTS process in general and the primary screening step in particular. There is a need for comprehensive statistical and computational data analysis systems that can characterize HTS data accurately and efficiently, including available prior information and \textit{borrowing information}. 

\section{HTS Data Structure}
\begin{figure}[h]
\centering
\includegraphics[scale=0.75]{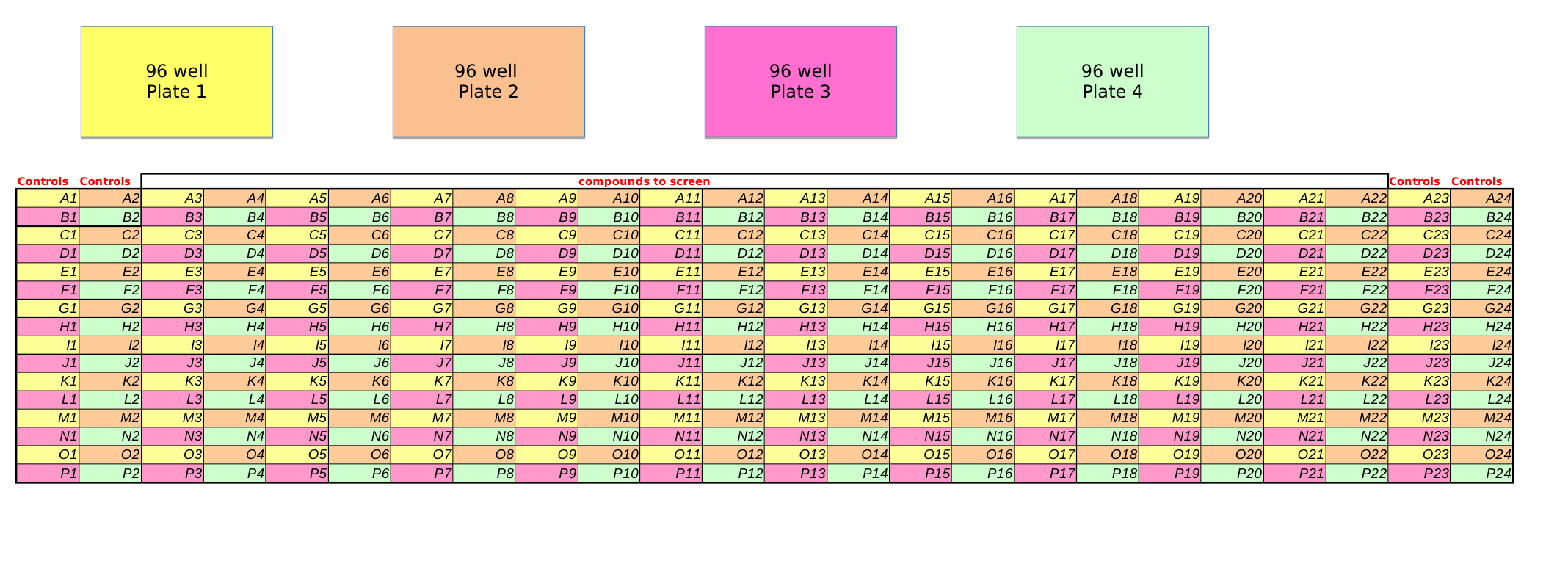}
\caption{384-well plate consisting of four 96-well plates. Figure taken from \citep{shterev1}.}
\label{384wellplate}
\end{figure}

Compounds are evaluated on 96-well or 384-well plates. In the example shown in Fig. \ref{384wellplate}, four 96-well plates are used to tile a 384-well plate. For each 96-well plate, the first and last columns typically contain only control wells, and thus a 96-well plate only contains $80$ test compounds. We assume that each well measures a different compound activity (e. g. no replicates) and has the same concentration of compound. It is also assumed that compounds are distributed randomly within the plate. The control wells are located in the first and last two columns of the 384-well plate. 

The 384-well plate design depicted in Fig. \ref{384wellplate} inherently creates cross-plate correlation among the individual 96-well plates. This type of correlation is not accounted for by simple HTS systems working on individual 96-well plates.

\begin{figure}[h]
\centering\scalebox{0.45}{
\includegraphics{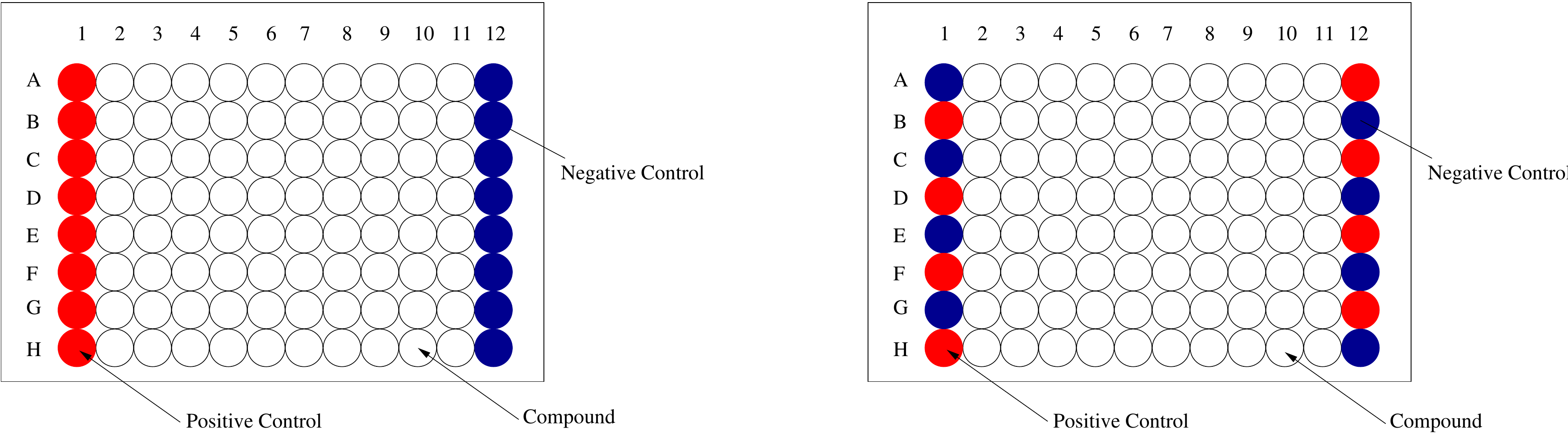}}
\caption{Example of a 96-well plate with compounds in the middle 80 wells and controls in the first and last column wells. Left panel shows a plate containing compounds, negative and positive controls. Right panel shows a 96-well plate in which positive and negative controls alternate to reduce plate edge effects.}
\label{96wellplate}
\end{figure}

Fig. \ref{96wellplate} provides a more detailed look of a 96-well plate. Ideally, controls should be placed randomly throughout the plate, to mitigate edge effects. However, the standard practice is to place the controls in the first and last columns and the compounds in inner columns. 

\section{Methods for High Throughput Screening of Compounds}
High throughput screening statistical practice \citep{malo,birmingham} has been focused on using simple methods such as the B-score, the Z-score and the normalized percent inhibition (NPI), for measuring compound activity and identifying potential candidate hits. These methods transform the compound raw value into the so called normalized value, which can then be used directly to assess compound activity. Each of the above mentioned methods has advantages and disadvantages and they differ in terms of how controls are used. The B-score and the Z-score do not use controls in the normalization process, while the NPI makes use of both positive and negative controls. 

The Z-score and the NPI work on per individual compound basis. The NPI, which has a biologically plausible interpretation as the percent activity relative to an established positive control, is defined as 
\begin{eqnarray}
NPI & = & \frac{z_p-z}{z_p-z_n}100\%,
\end{eqnarray}
where $z$ is the compound raw value and $z_n$ and $z_p$ are the negative and positive control raw values respectively.

The Z-score is defined as
\begin{eqnarray}
Z & = & \frac{z-\mu_z}{\sigma_z},
\end{eqnarray}
where $\mu_z$ and $\sigma_z$ are the mean and standard deviation respectively of all compounds in the plate.

The B-score works on a per plate basis in the sense that the plate geometry has an effect on the computed score. The B-score is defined as
\begin{eqnarray}
B & = & \frac{r_z}{MAD_z},
\end{eqnarray}
where $r_z$ is a matrix of residuals obtained after a median Polish fitting procedure and $MAD_z$ is the median absolute deviation.  

\begin{figure}[h]
\centering
\includegraphics[scale=0.6]{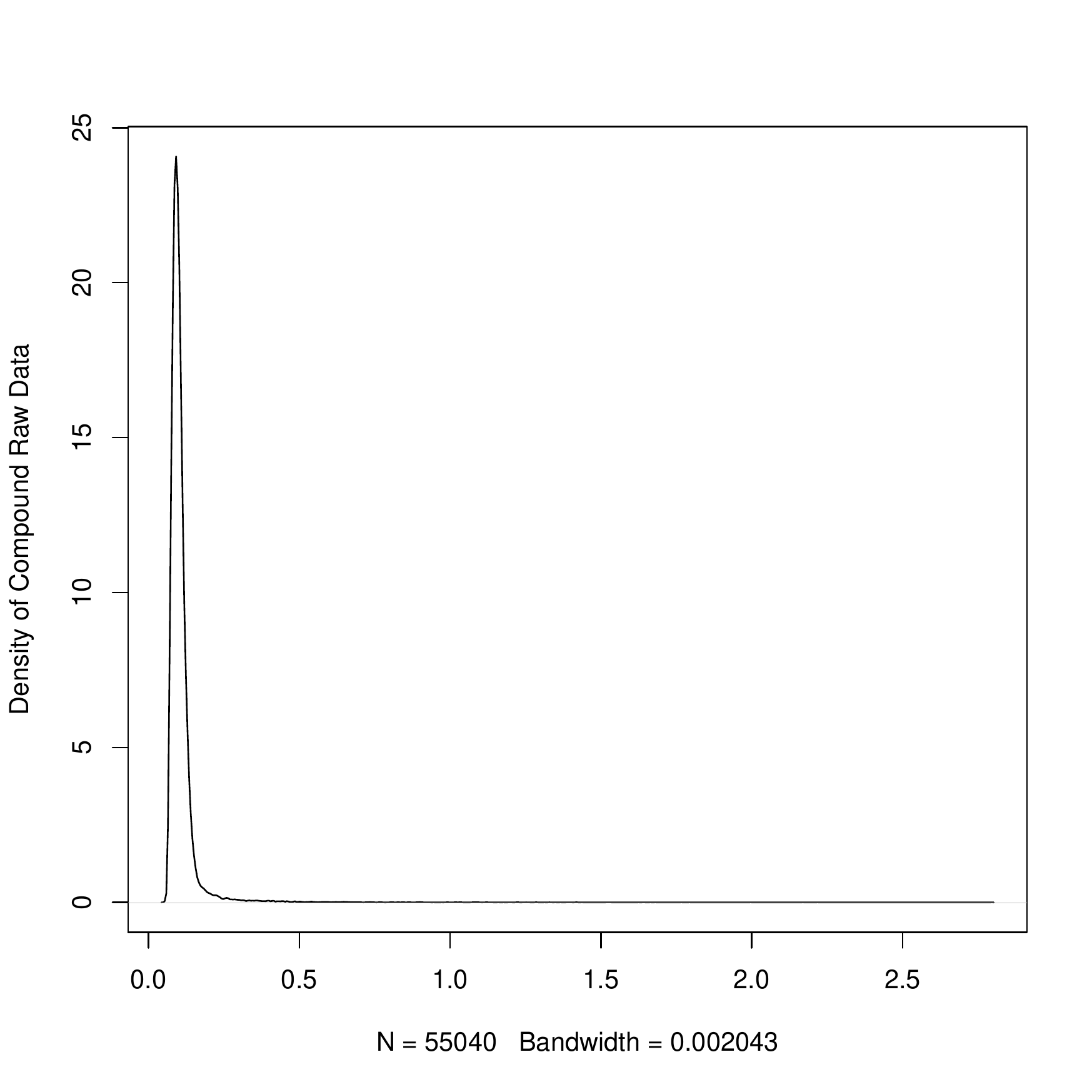}
\caption{Density of mast cell activated compounds, exhibiting a log-normal law with a large positive outlier. Data are under the auspices of NIH contract No. HHSN272201400054C.}
\label{densityOD}
\end{figure}

The NPI, Z-score and B-score all have significant limitations. The NPI is very sensitive to edge effects, since it uses the negative and positive control wells that are typically in the outer columns. The Z-score is susceptible to outliers and assumes normally distributed compound readout values, an implausible assumption in many screening contexts (Fig. \ref{densityOD}). Although the B-score takes into account systematic row and column plate effects and is the method of choice \citep{malo} in many cases, it requires an arbitrary threshold to identify hits and tends to miss important compounds with minimal or moderate activity. Critically, all these methods treat each plate independently. In some cases, systematic experimental and plate design effects may induce correlation among groups of plates. It is therefore desirable to have a system that works on multiple plates simultaneously.  

A Bayesian approach for hit selection in RNAi screens was proposed in \citep{zhang}. The model imposes separate Gaussian priors on active, inactive and inhibition siRNAs. Inference is based on hypothesis testing via posterior distributions. The posterior distributions are then directly used to control false discovery rate (FDR) \citep{newton}. The proposed method is parametric and although it may be reasonable in some cases to model the siRNAs as normally distributed, many data in practice and particularly HTS data are not Gaussian (as shown in Fig. \ref{densityOD}). Additionally, the priors of the proposed method incorporate common information that is pooled from all plates. This type of information sharing is fixed and is different from the multi-plate sharing mechanism in machine learning, where different groups of data iteratively and selectively share information via a global layer \citep{teh}.

In this paper we develop a new system for HTS of compounds based on Bayesian statistics. The nonparametric method does not use controls and is capable of characterizing HTS data that are not necessarily Gaussian distributed. It can handle multiple plates simultaneously and is able to selectively share statistical strength among plates. This selective sharing mechanism, that is being updated at each sampler iteration, is important for discovering systematic experimental effects that propagate differently among plates. We develop an efficient Markov chain Monte Carlo (MCMC) sampler for estimating the compound readout posteriors. Based on posterior probabilities specifying if a compound is active or not, it is possible to determine probabilistic significance, control FDR \citep{whittemore} and adjust for multiple comparison \citep{scott} in a Bayesian hierarchical manner. The framework is implemented as an {\tt R} extension package {\tt BHTSpack} \citep{shterev2}.

\section{Statistical Model}
Dirichlet process Gaussian mixtures (DPGM) \citep{antoniak, escobar} constitute a powerful class of models that can nonparametrically describe a wide range of distributions encountered in practice. The simplicity of DPGM and their ease of implementation have made them a preferable choice in many applications, as well as building blocks of more complex models and systems. In the DPGM framework, the Dirichlet process (DP) \citep{ferguson, sethuraman} models the mixing proportions of the Gaussian components. The hierarchical Dirichlet process (HDP) \citep{teh} is particularly suitable for modeling multi-task problems in machine learning. These problems seem to be very analogous to our multi-plate HTS of compounds scenario. 

Our framework deploys two HDPs to characterize the active and inactive components. In the following sequel we approximate the DP via the finite stick-breaking representation \citep{ishwaran}. Let $m\in\{1,\ldots,M\}$ denote the plate index, where $M$ is the total number of plates. Let $i\in\{1,\ldots,n_m\}$ denote the compound well index within a plate, $h\in\{1,\ldots,H\}$ denote the DP mixture cluster index within a plate and $k\in\{1,\ldots,K\}$ denote the global DP component index. Let superscripts $(1)$ and $(0)$ refer to active and inactive compounds, respectively. Motivated in part by \citep{lock}, we propose the following hierarchical fully Bayesian HTS (BHTS) model:
\begin{eqnarray}
z_{mi} & \sim & \pi\sum_{h=1}^H\lambda_{mh}^{(1)}\mathcal{K}(z_{mi};\theta_{h}^{(1)})+(1-\pi)\sum_{h=1}^H\lambda_{mh}^{(0)}\mathcal{K}(z_{mi};\theta_{h}^{(0)})\\
\pi & \sim & \mbox{Beta}(a_\pi, b_\pi)\\
(\lambda_{m1}^{(1)},\ldots,\lambda_{mH}^{(1)})=G_{m}^{(1)} & \sim & \mbox{DP}\big(\alpha_1,\boldsymbol{\lambda_H^{(1)}\big)}\\%\mbox{Dirichlet}(\alpha_1\lambda_{1}^{(1)},\ldots,\alpha_1\lambda_{H}^{(1)})\\
(\lambda_{m1}^{(0)},\ldots,\lambda_{mH}^{(0)})=G_{m}^{(0)} & \sim & \mbox{DP}\big(\alpha_0,\boldsymbol{\lambda_H^{(0)}\big)}\\%\mbox{Dirichlet}(\alpha_0\lambda_{1}^{(0)},\ldots,\alpha_0\lambda_{H}^{(0)})\\
\alpha_1,\, \alpha_0 & \sim & \mbox{Ga}(a_{\alpha}, b_{\alpha})\\
(\lambda_{1}^{(1)},\ldots,\lambda_{K}^{(1)})=G_{0}^{(1)} & \sim & \mbox{DP}\big(\tau_1,\boldsymbol{\lambda_K^{(1)}\big)}\\%\mbox{Dirichlet}(\tau_1/K,\ldots,\tau_1/K)\\
(\lambda_{1}^{(0)},\ldots,\lambda_{K}^{(0)})=G_{0}^{(0)} & \sim & \mbox{DP}\big(\tau_0,\boldsymbol{\lambda_K^{(0)}\big)}\\%\mbox{Dirichlet}(\tau_0/K,\ldots,\tau_0/K)\\
\tau_1,\, \tau_0 & \sim & \mbox{Ga}(a_{\tau}, b_{\tau})\\
\theta_{h}^{(1)} & \sim & \mathcal{N}(\mu_1|\mu_{10}, \sigma_1^2)\mbox{Inv-Ga}(\sigma_1^2|a,b)\\
\theta_{h}^{(0)} & \sim & \mathcal{N}(\mu_0|\mu_{00}, \sigma_0^2)\mbox{Inv-Ga}(\sigma_0^2|a,b)
\end{eqnarray}
where $\mathcal{K}(\cdot;\theta)$ denotes a Gaussian kernel with parameters $\theta$, and $\mbox{DP}(\cdot,\cdot)$ denotes the finite stick-breaking representation of the DP (see supplementary materials for more details).

\begin{figure}[htbp]
\centering\scalebox{0.6}{
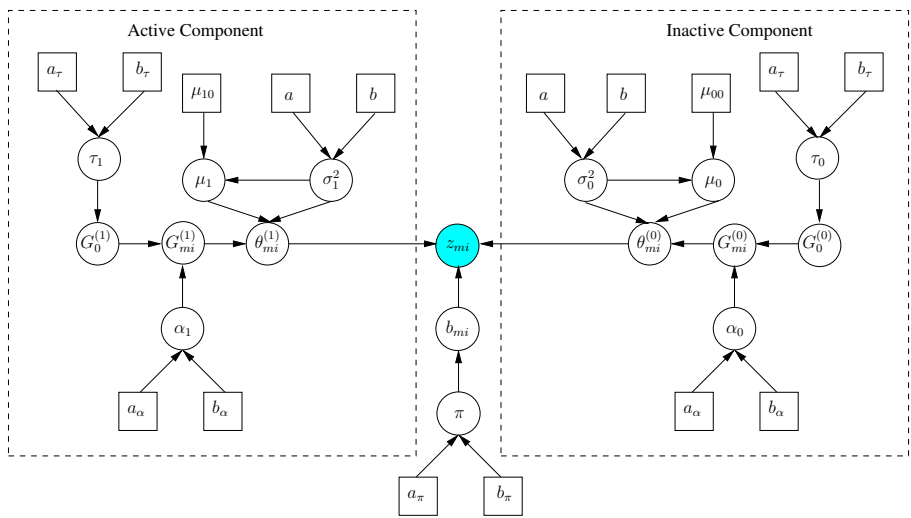}
\caption{A graphical 
representation of BHTS model. The latent binary variable $b_{mi}$ specifies active ($1$) or inactive ($0$) compound.}
\label{fig:graphmodel}
\end{figure}

A graphical representation of the BHTS model is shown in Fig. \ref{fig:graphmodel}. The blue circle represents the observed variable, white circles represent hidden (latent) variables and squares represent hyper-parameters. Conditional dependence between variables is shown via the directed edges.

The compound data mean and variance can be used to specify the model hyperparameters, without the help from controls. A compound can be \emph{active} (potential candidate hit) or \emph{inactive} (exhibiting no activity). 

Prior information about active and inactive compounds is reflected in the model hyperparameters $\mu_{10}$ and $\mu_{00}$, respectively. These hyperparameters specify compound activity level estimates and can be specified using the mean of the compound data. For example, $\mu_{10}$ and $\mu_{00}$ can be set somewhat larger and smaller than the compound data mean, respectively. 

The compound variability hyperparameters $\{a,b\}$ are common for both active and inactive compounds, but the model facilitates different active $\sigma_1^2$ and inactive $\sigma_0^2$ compound variabilities. Let $v$ denote the compound data variance, which can be used to specify $\{a,b\}$. Setting the inverse gamma variance to $10^{-4}$ for example, and using $v$ as its mean, the inverse gamma density parameters can be derived as:

\begin{eqnarray}
a & = & \frac{v^2}{10^{-4}} + 2,\\
b & = & \frac{v^3}{10^{-4}} + v.
\end{eqnarray}

We derive MCMC update equations based on approximate full conditional posteriors of the model parameters and construct a Gibbs sampler that iteratively samples from these update equations. The stick breaking construction \citep{ishwaran} is used in the approximate posterior distributions of the global and local DP weights. The update equations are shown in the supplementary materials.

\section{False Discovery Rate and Multiplicity Correction}
Our problem can be formulated as performing $\sum_{m}n_m$ dependent hypothesis tests of $b_{mi}=0$ versus $b_{mi}=1$. %A local FDR at point $z_{mi}$ \citep{efron} can be computed as:
%\begin{eqnarray}
%\mbox{fdr}(z_{mi}) & = & 1-\hat{\pi}(z_{mi}),
%\end{eqnarray} 
%where $\hat{\pi}(z_{mi})$ is the posterior probability estimate of the compound $z_{mi}$ being a hit (see supplementary). 
Following \citep{muller}, an estimate to FDR for a given threshold $r$ can be computed as:
\begin{eqnarray}
\overline{\mbox{FDR}}(r) & = & \frac{\sum_{m,i}1\big(\hat{\pi}(z_{mi})>r\big)\big(1-\hat{\pi}(z_{mi})\big)}{\sum_{m,i}1\big(\hat{\pi}(z_{mi})>r\big)},
\end{eqnarray}
where $\hat{\pi}(z_{mi})$ is the posterior probability estimate of the compound $z_{mi}$ being a hit (see supplementary) and $1(\cdot)$ is the indicator function.
%Assume that for an index $t\in\{1,\cdots,\sum_{m}n_m\}$, the probabilities $\hat{\pi}(z_{t})$ are in increasing order. The FDR can be controlled at a given level $\gamma$ via a Benjamini and Hochberg type rule \citep{muller} that rejects all hypothesis for which:
%\begin{eqnarray}
%\mbox{fdr}(z_{t}) & \leq & \frac{t\gamma}{\sum_{m}n_m}.
%\end{eqnarray}
Since the model is fully Bayesian, multiple comparison is automatically accounted for \citep{scott} in the estimated posteriors $\hat{\pi}(z_{mi})$.

\section{Test Data}
We assess the performance of the BHTS method using a synthetically generated data set and compare it with the B-score method in terms of receiver operating characteristic (ROC) curves and area under the curve (AUC). The {\tt R} extension package {\tt pROC} \citep{xavier} was employed in the analysis.

We also perform experiments with data sets containing real chemical compounds and controls with low, medium and high activity. These real data set experiments are used to assess the proposed method capability to identify hits with low and medium activity levels. A comparison with the industry standard B-score method is provided. 

\subsection{Synthetic Compound Data}
We constructed synthetic data for the purpose of assessing sensitivity and specificity of the proposed algorithm. We generated a set of $80\times10^3$ compounds consisting of hits and non-hits. The hits were generated from a four component log-normal mixture model with means $\{0.20, 0.24, 0.28, 0.32\}$ and variances $\{0.0020, 0.0022, 0.0024, 0.0026\}$. Similarly, the non-hits were generated from a four component log-normal distribution with means $\{0.10, 0.12, 0.14, 0.16\}$ and variances $\{0.010, 0.011, 0.012, 0.013\}$. The compounds were then randomly distributed among $1000$ compound plates, with each plate consisting of eight rows and ten columns.

\begin{figure}[h]
\centering
\includegraphics[scale=0.3]{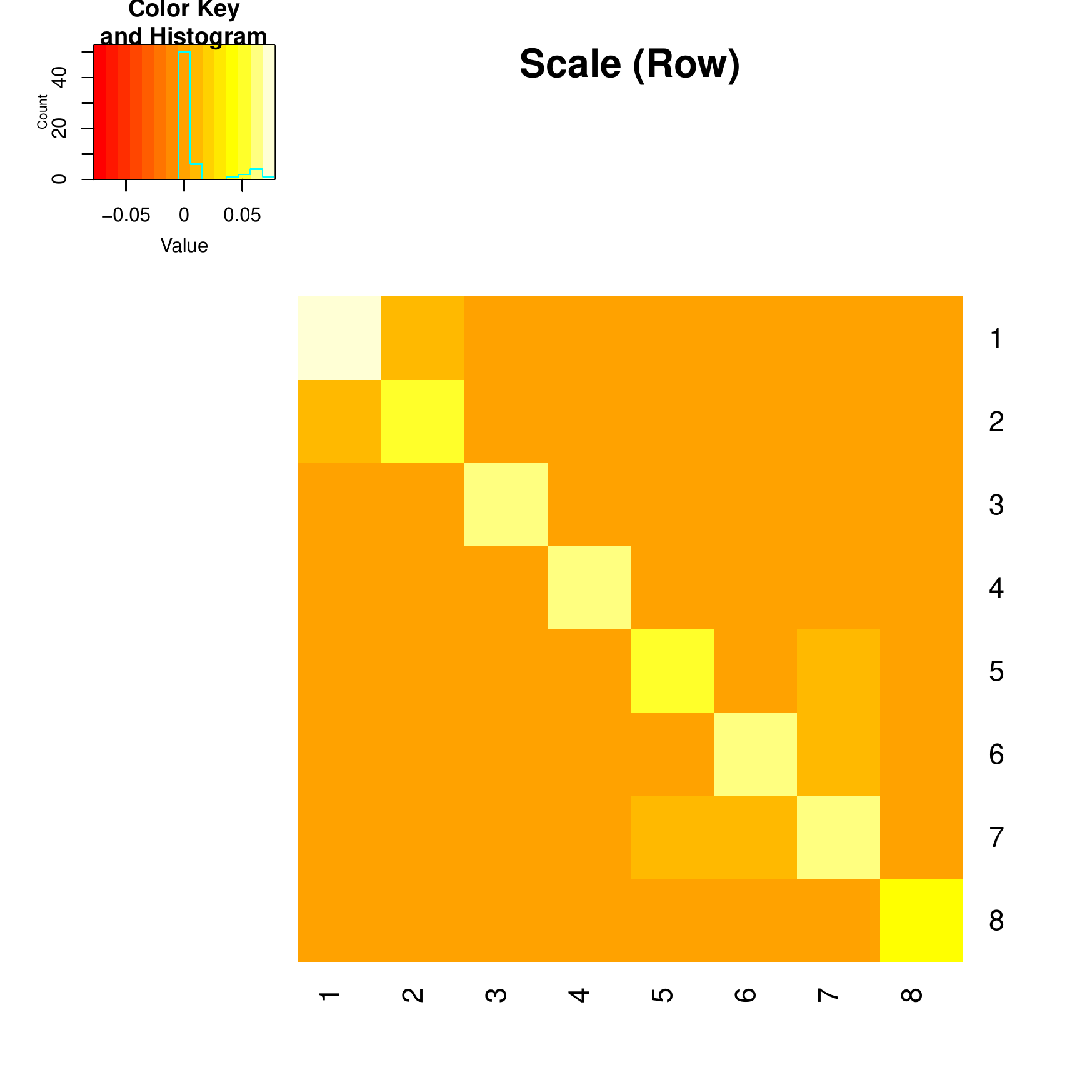}
\includegraphics[scale=0.3]{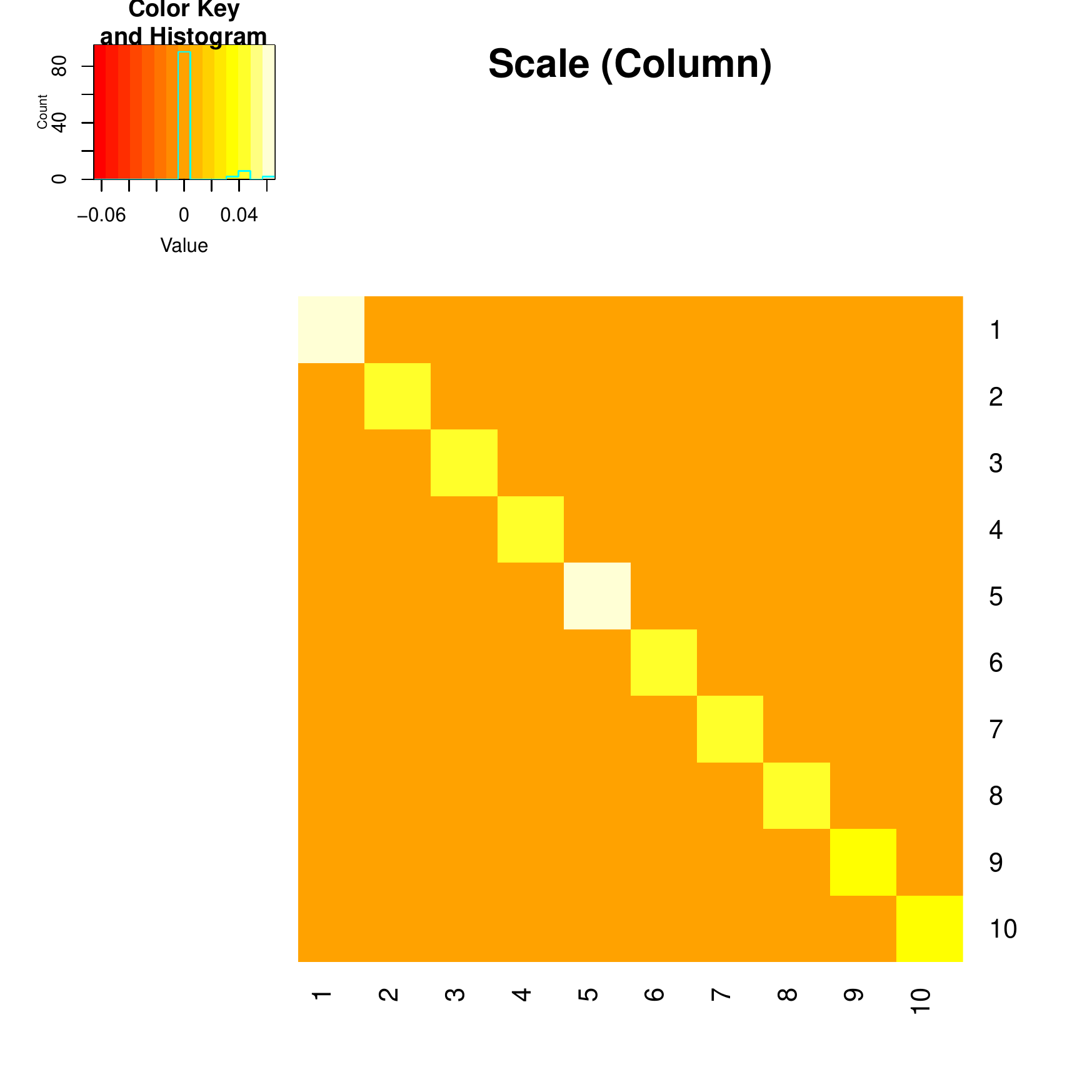}
\includegraphics[scale=0.35]{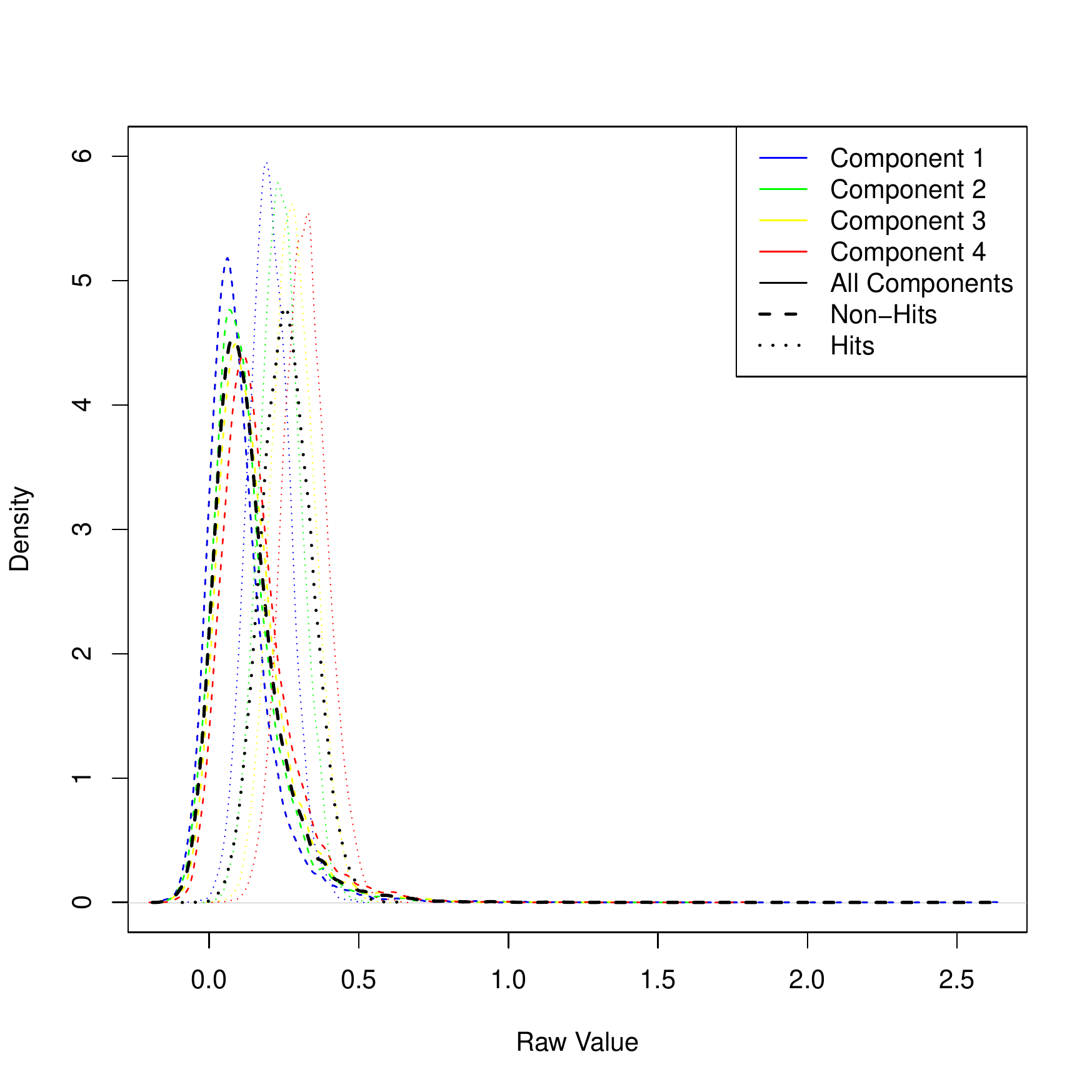}
\caption{Synthetic compound data used in the experiments. Left and middle plots show scale matrices used in generating the synthetic noise plates, indicating predominantly row dependent within plate effects. Right plot shows density of resulting synthetic compound and plate effect data.}
\label{fig:scale}
\end{figure}

We simulated plate effects by generating random noise from the matrix-normal distribution, with a zero location matrix and specific row and column scale matrices. The row and column scale matrices were designed in a way to reflect the structure of within plate row and column effects encountered in practice. Specifically, we used a real data set of compounds exhibiting the plate design shown in Fig. \ref{384wellplate}. We excluded all control well columns and computed B-scores based on individual $8\times 10$ compound well plates. We then estimated the row-wise and column-wise covariance matrices of the difference between the compound raw values and their B-scores. The estimated covariance matrices were then properly scaled and used as row and column scale matrices (shown in Fig. \ref{fig:scale}) in generating the plate noise effects. An independently drawn noise plate was added to each of the compound plates. The resulting data plates were used as test data.

Experiments were performed with data sets containing different proportions of active and inactive compounds. Considering the fact that a large collection of compounds will probably contain a relatively small number of candidate compound hits of interest, we experimented with data sets containing $40\%$, $10\%$, and $5\%$ of active compounds, respectively. In all synthetic data set experiments, the model hyperparameters were the same. See supplementary for specific hyperparameter values.

\clearpage
\subsubsection{Comparison with B-score}
\begin{figure}[h]
\centering
\includegraphics[scale=0.3]{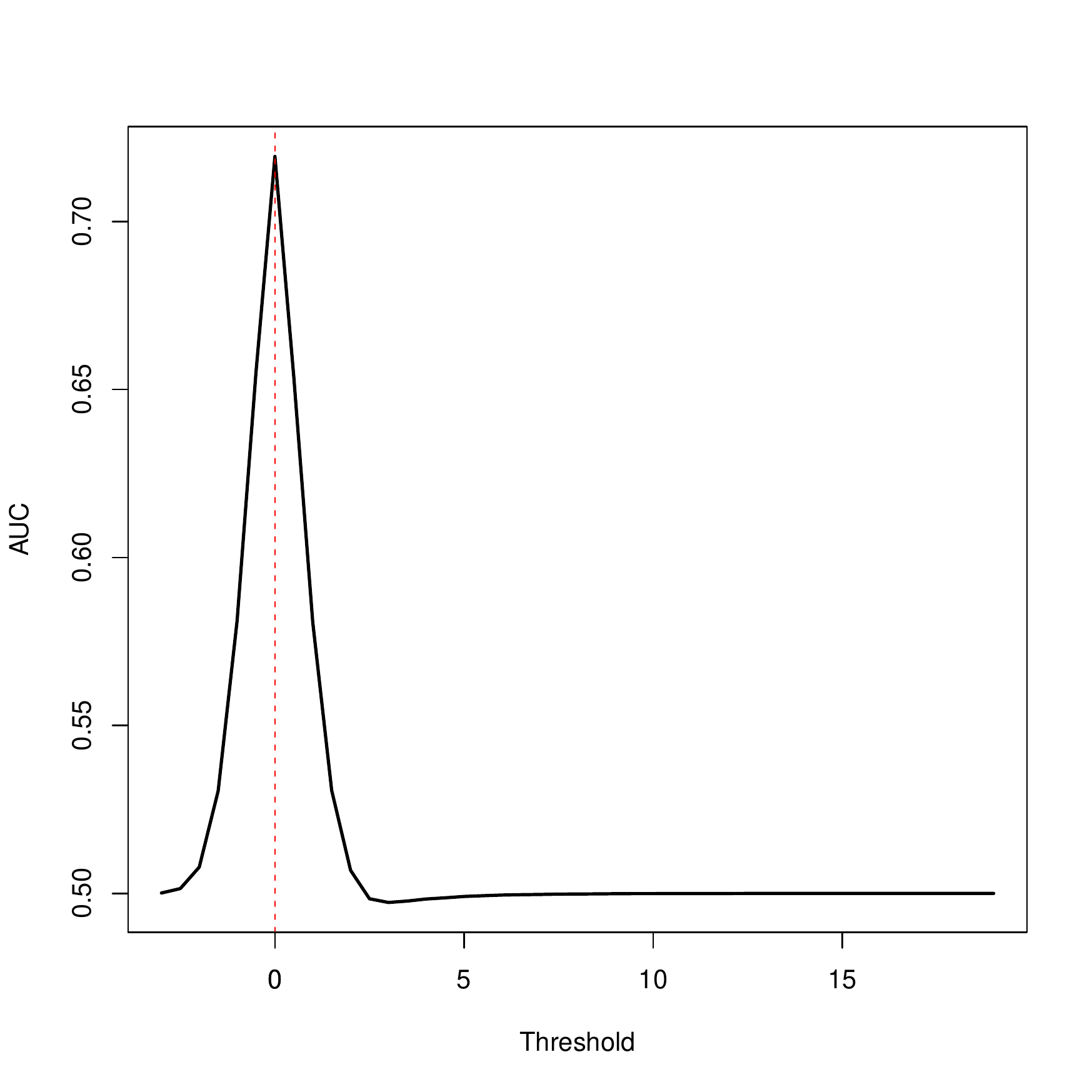}
\includegraphics[scale=0.3]{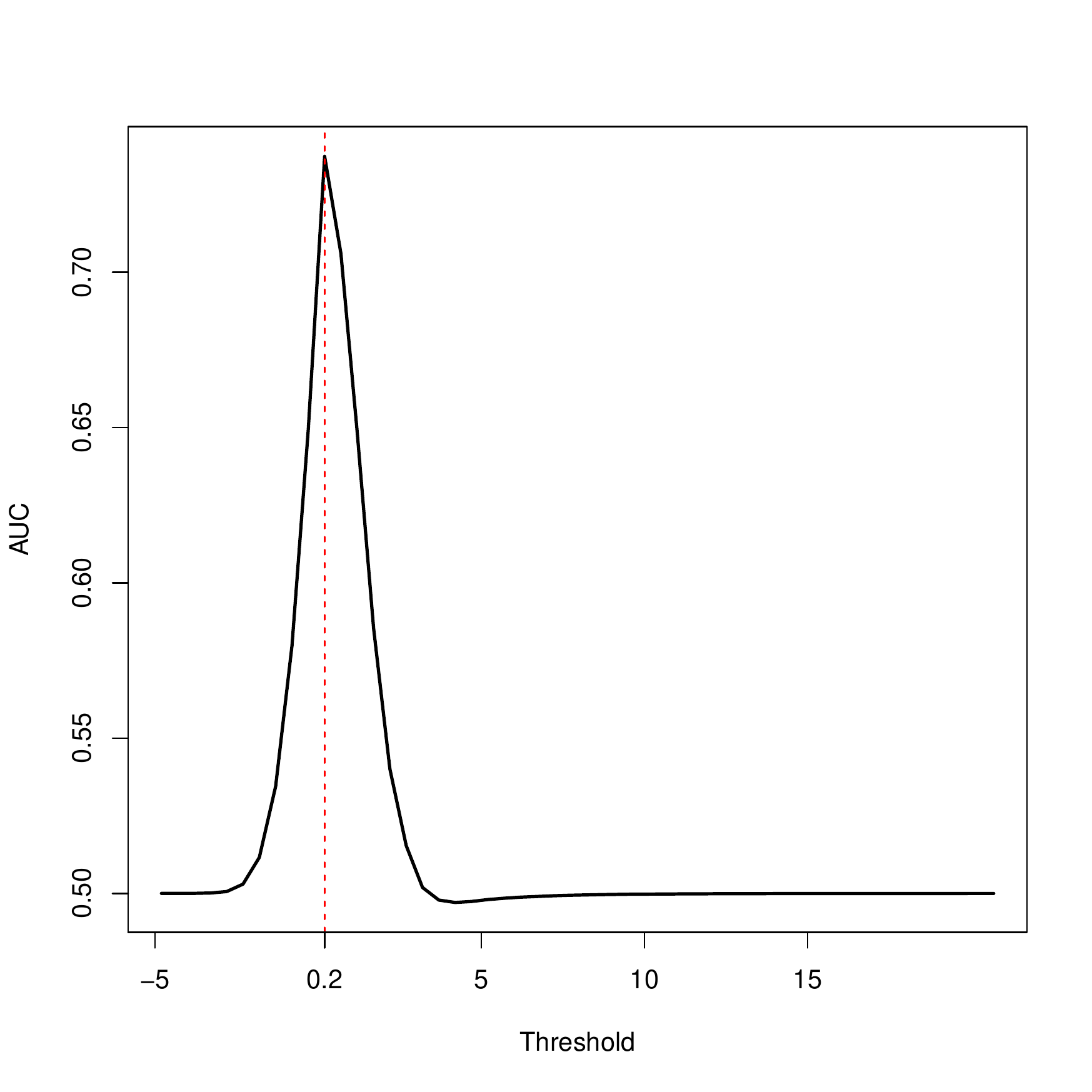}
\includegraphics[scale=0.3]{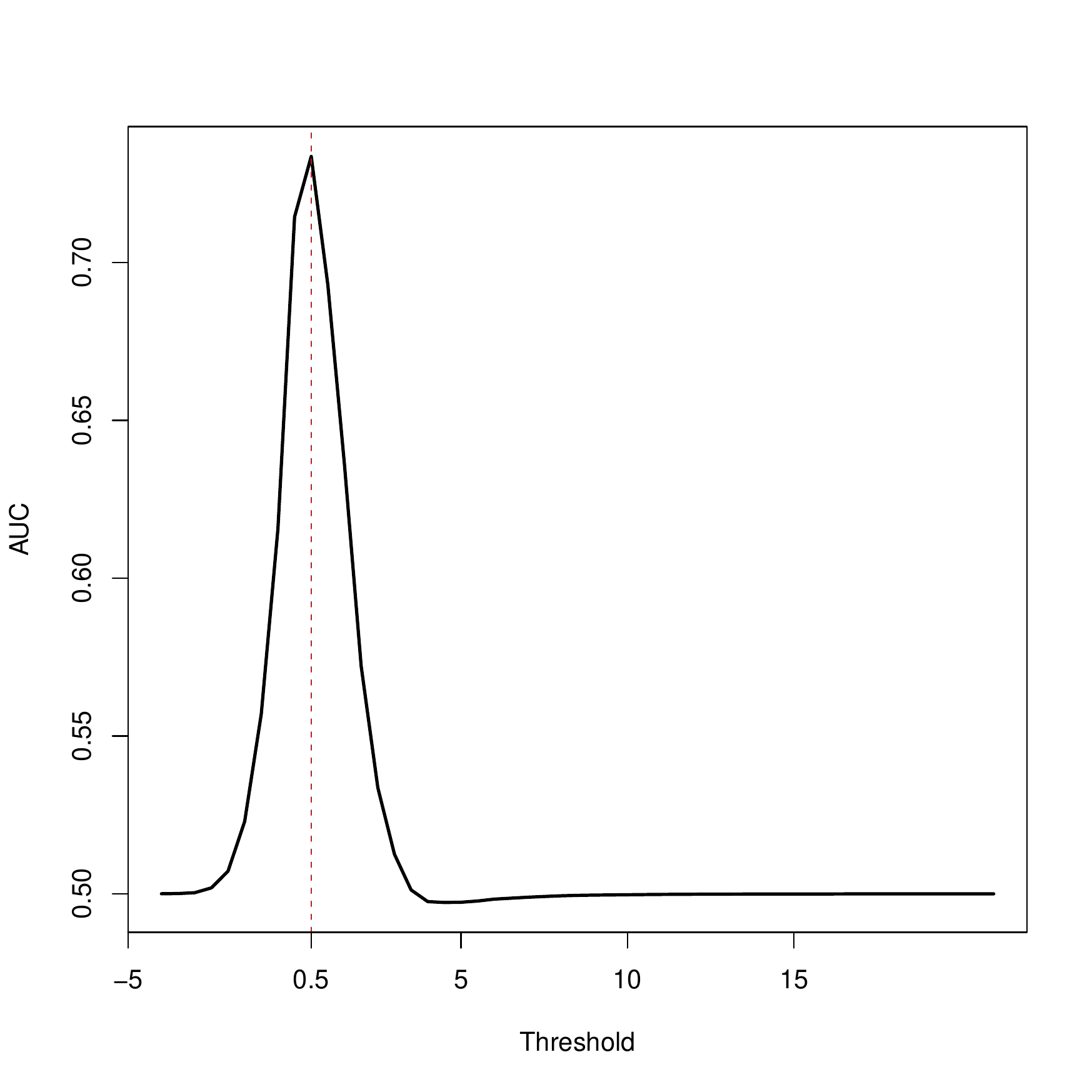}
\includegraphics[scale=0.3]{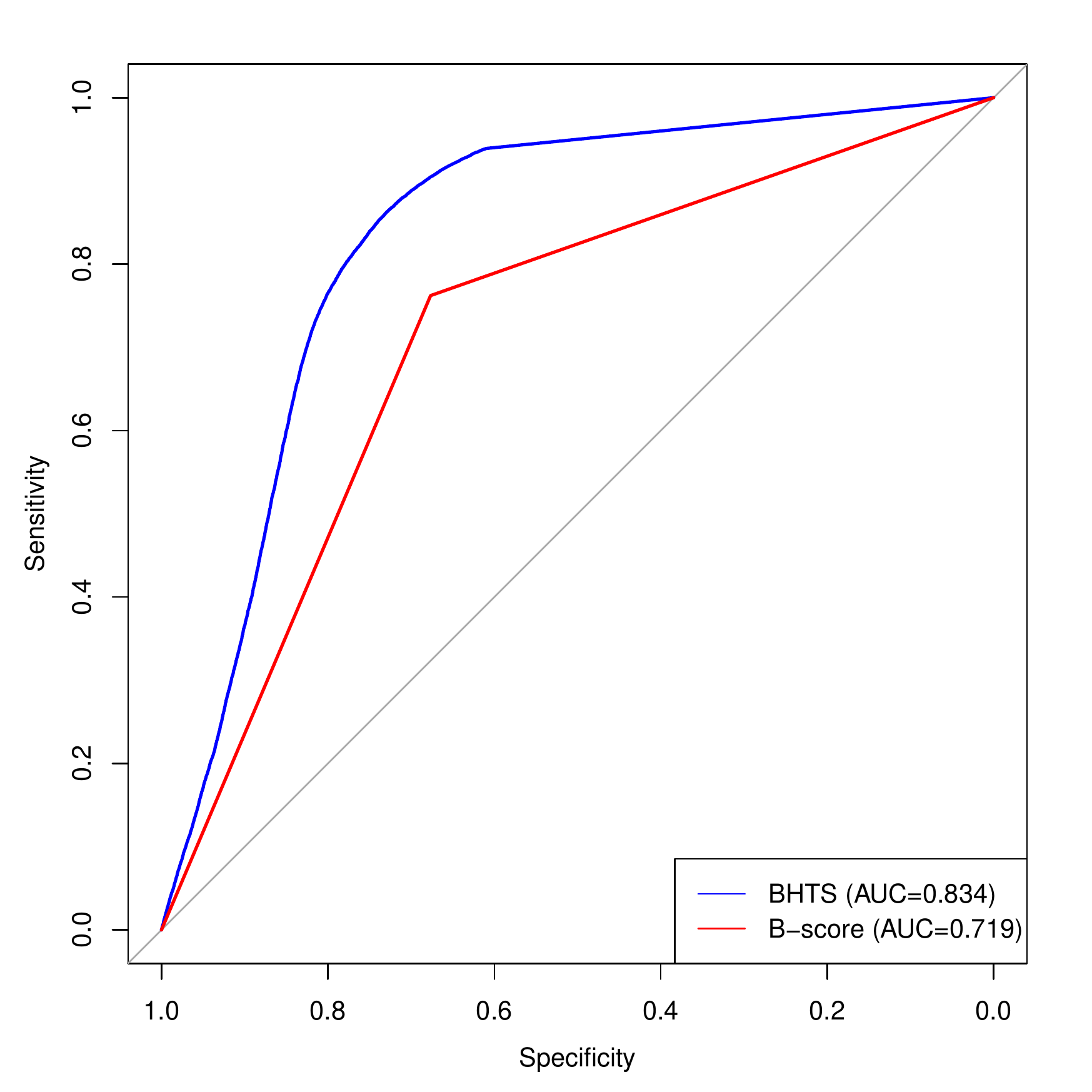}
\includegraphics[scale=0.3]{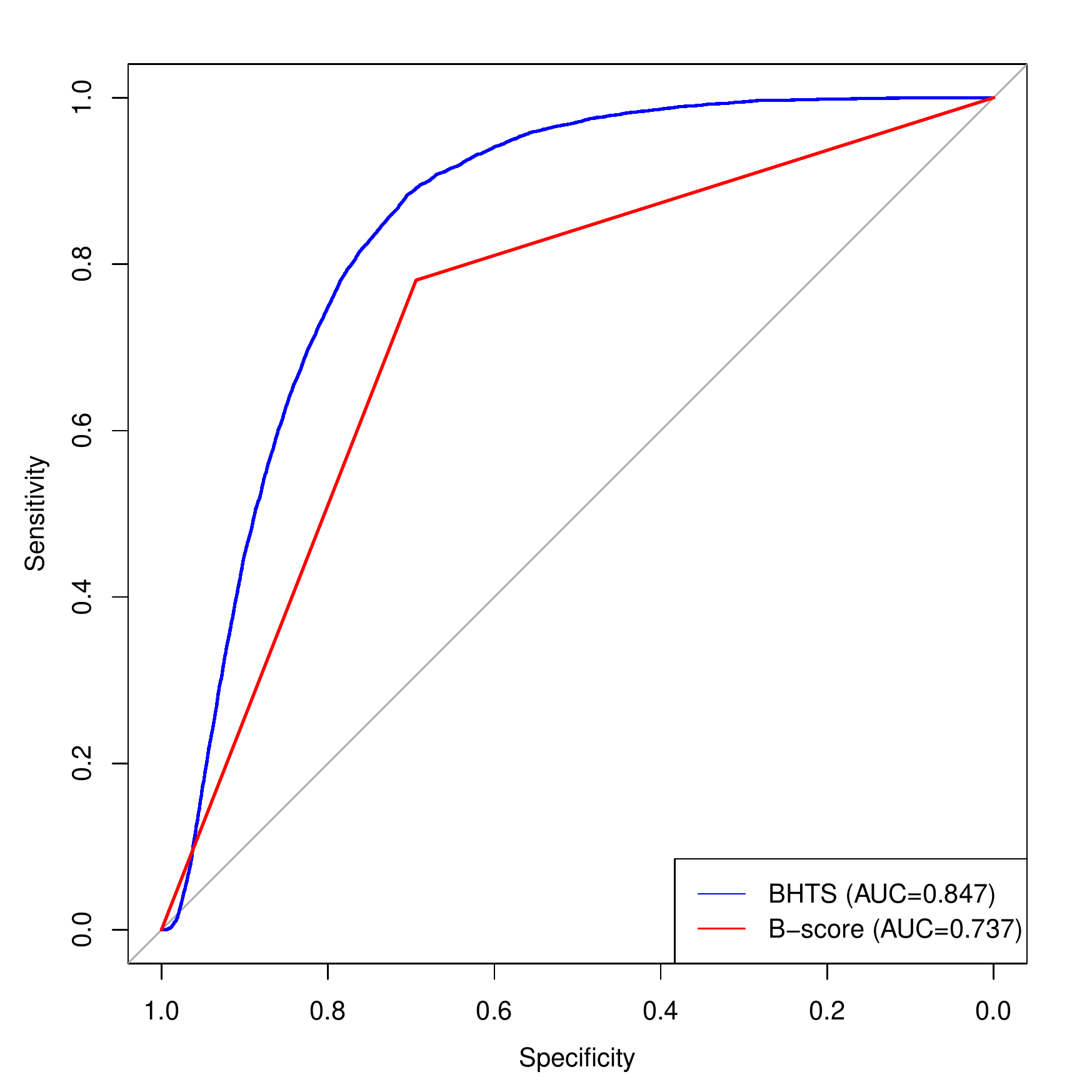}
\includegraphics[scale=0.3]{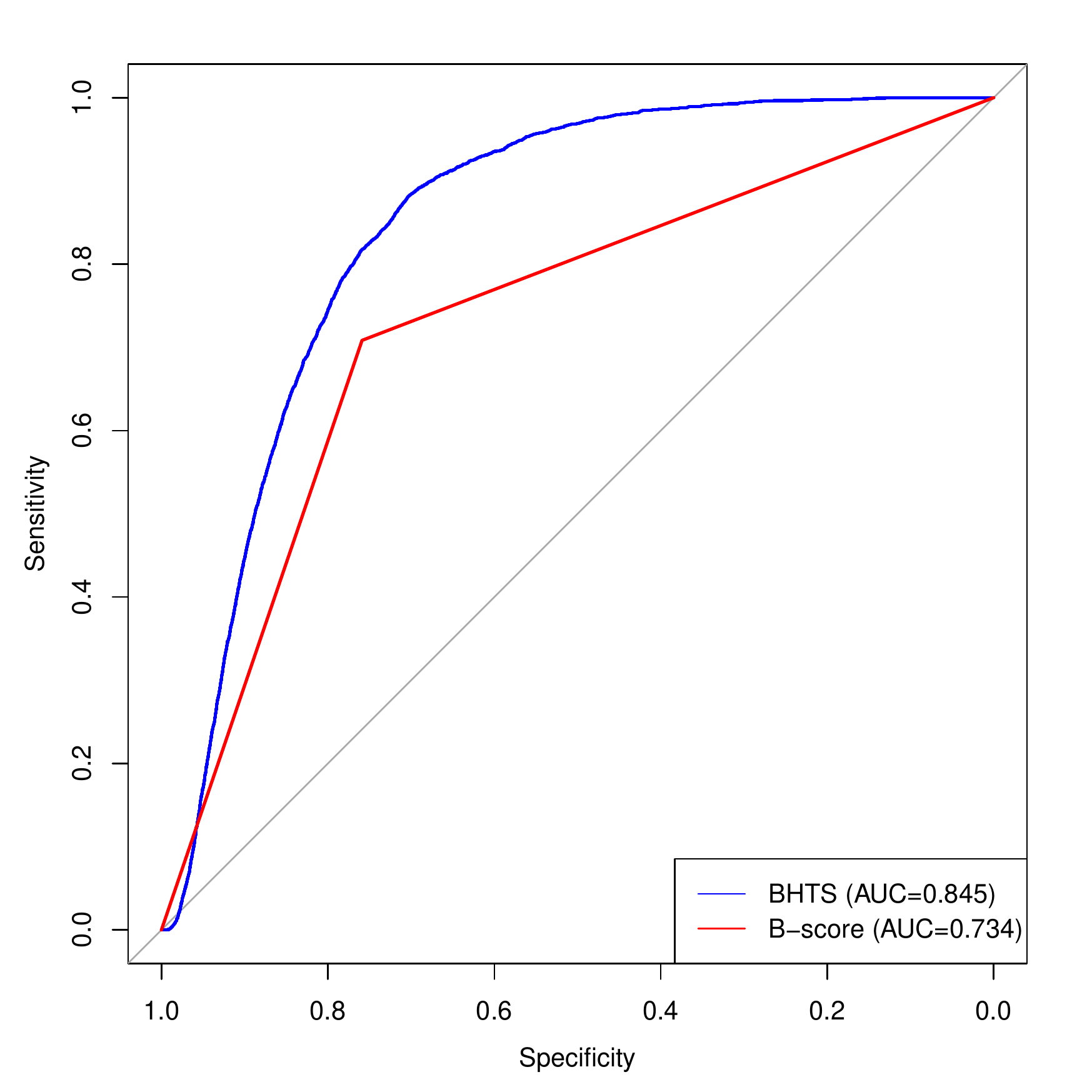}
\caption{Top row shows B-score method AUC plots as functions of thresholds. Bottom row shows ROC plots of the B-score and BHTS methods. Data sets containing $40\%$ (left column), $10\%$ (middle column) and $5\%$ (right column) of active compounds, respectively.}
\label{fig:auc_roc}
\end{figure}

Experimental ROC results are shown in Fig. \ref{fig:auc_roc}. The B-score ROC curve is based on the maximum achievable AUC threshold shown in the same figure. It can be seen that the BHTS method improves upon the B-score method in terms of classification accuracy. The results in Fig. \ref{fig:auc_roc} also demonstrate that the B-score is highly sensitive to a particularly chosen optimal threshold, as evidenced by the spike in the AUC curve as a function of the threshold.

\subsubsection{Hyperparameter Sensitivity Analysis}
\begin{figure}[h]
\centering
\includegraphics[scale=0.3]{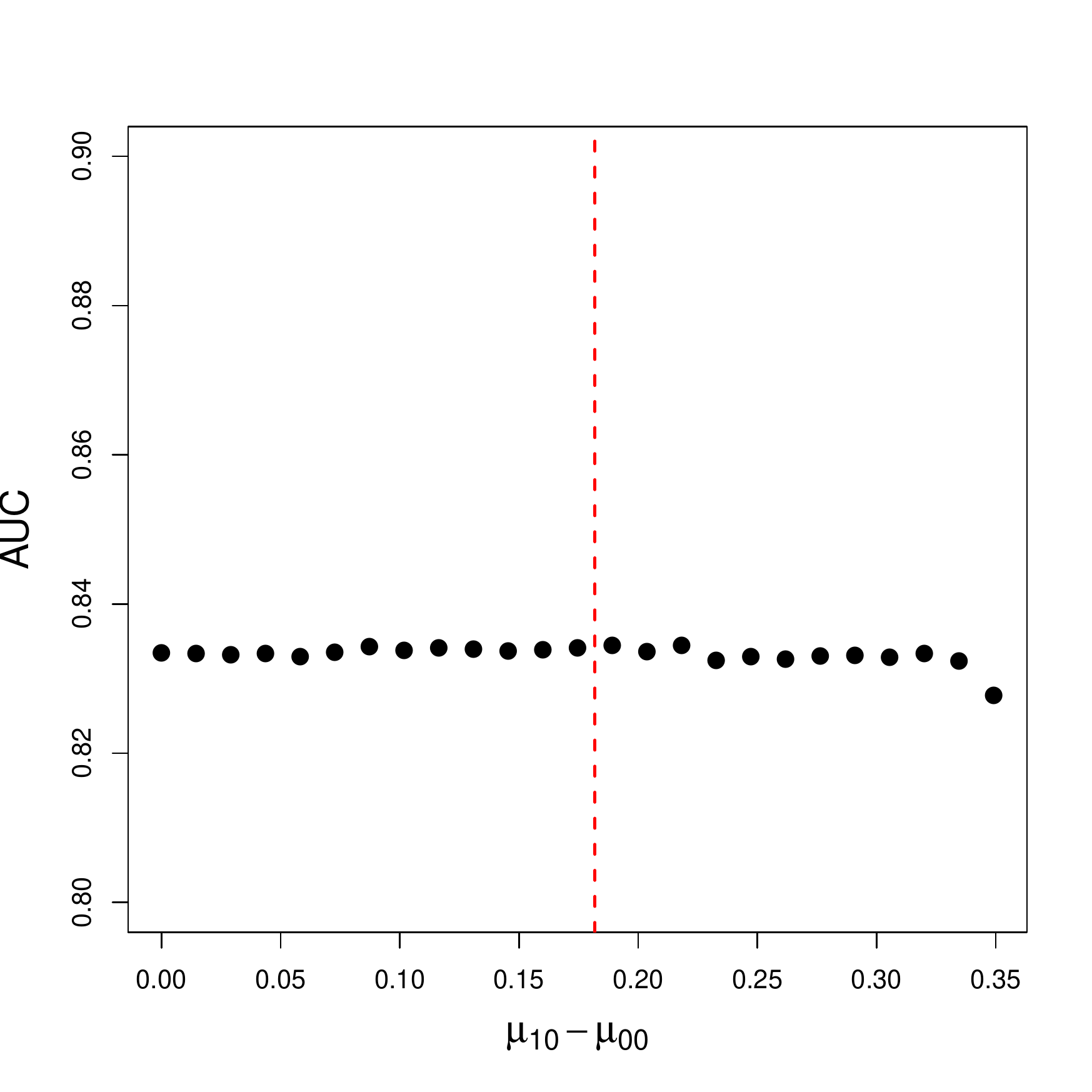}
\includegraphics[scale=0.3]{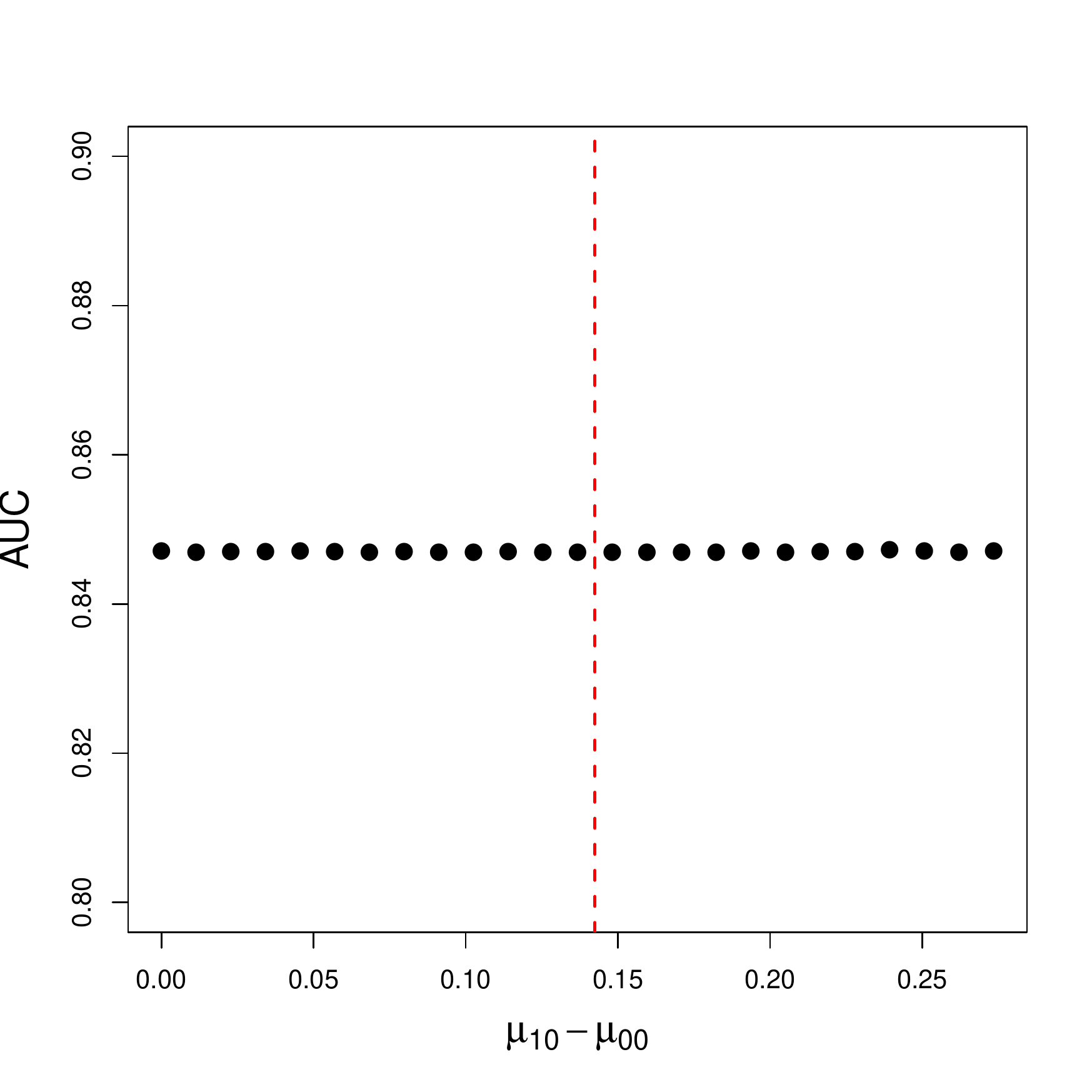}
\includegraphics[scale=0.3]{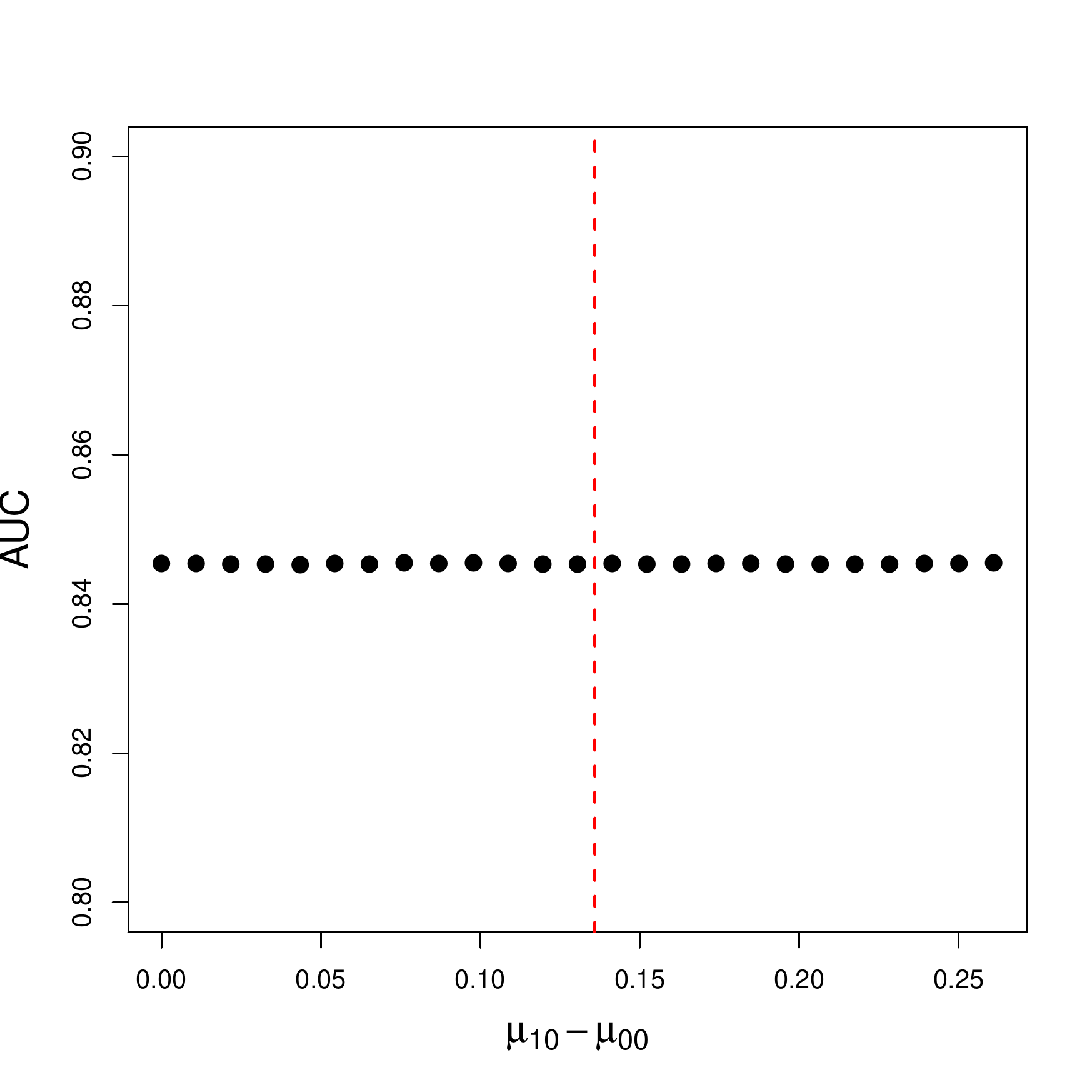}
\caption{AUC as a function of $(\mu_{10}-\mu_{00})$, for data sets containing $40\%$ (left plot), $10\%$ (middle plot) and $5\%$ (right plot) of active compounds. Red line indicates mean of compound data.}
\label{fig:sensitivity}
\end{figure}

In this subsection we assess the sensitivity of the proposed method to the choice of hyperparameter values $\{\mu_{10}, \mu_{00}\}$ by computing the AUC for a range of values of the difference $(\mu_{10}-\mu_{00})$. Experimental results are shown in Fig. \ref{fig:sensitivity}. It can be seen that the model performs similarly in terms of AUC, for a range of $(\mu_{10}-\mu_{00})$ values and different data sets. 

\subsection{Real Compound Data}
\begin{figure}[h]
\centering
\includegraphics[scale=0.32]{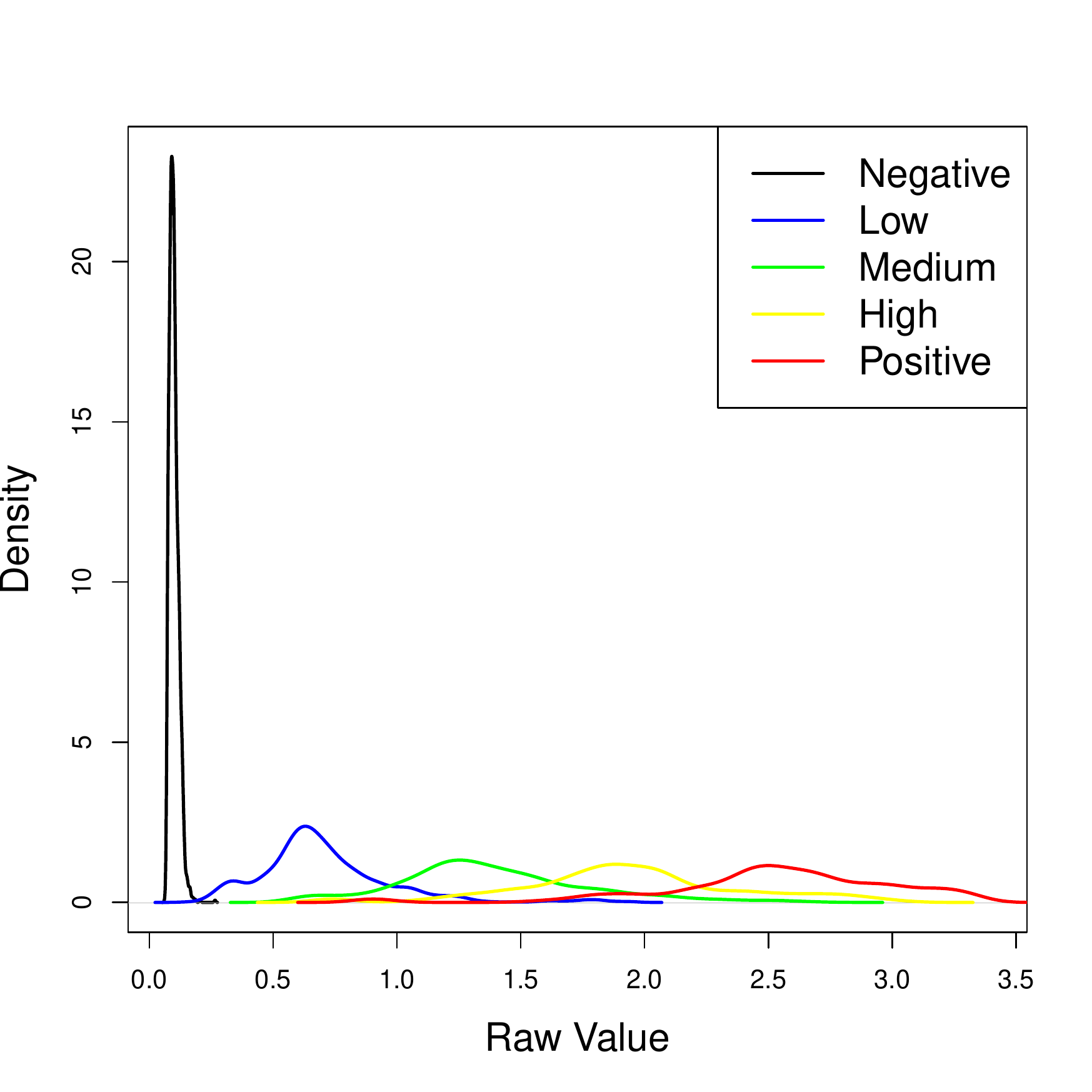}
\includegraphics[scale=0.32]{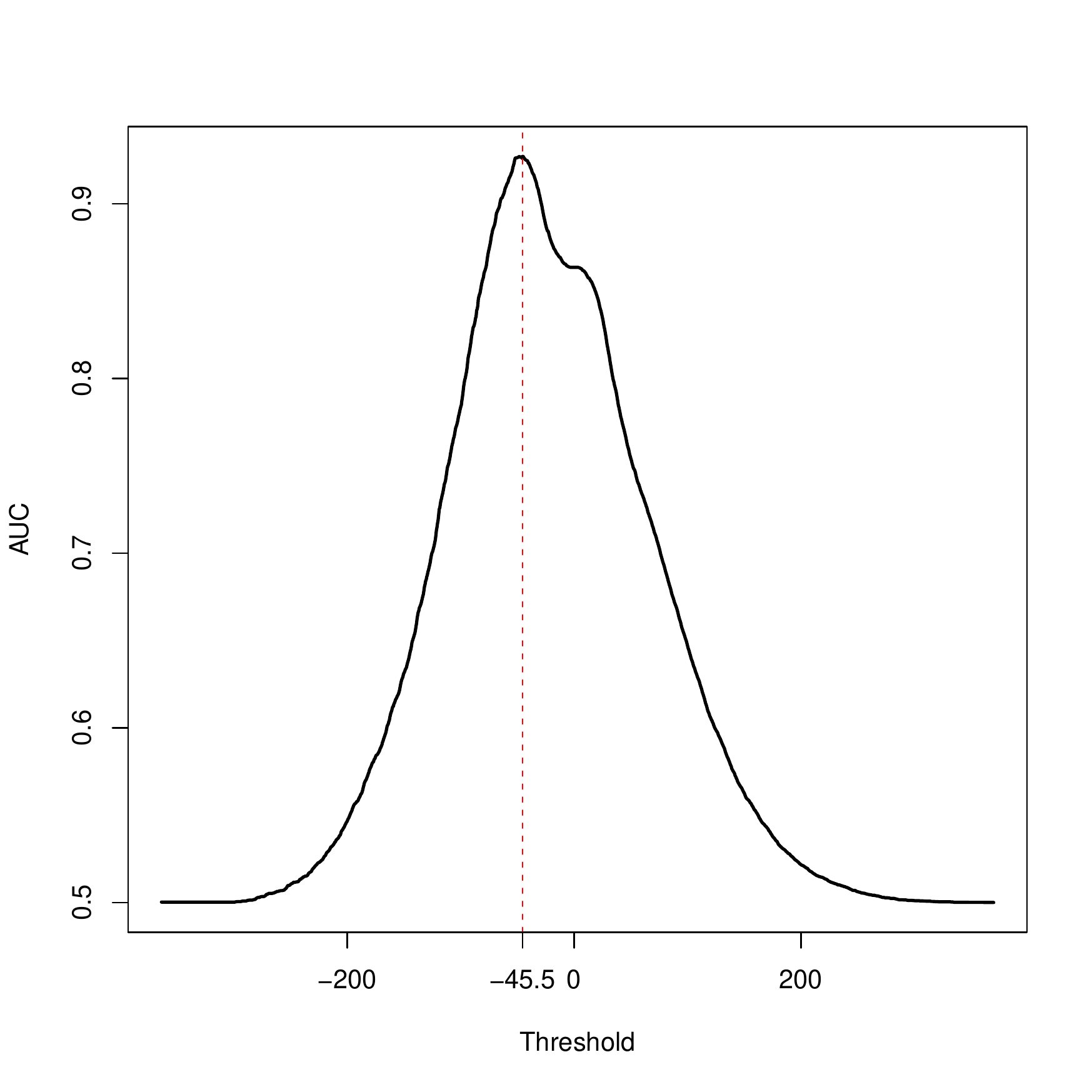}
\includegraphics[scale=0.3]{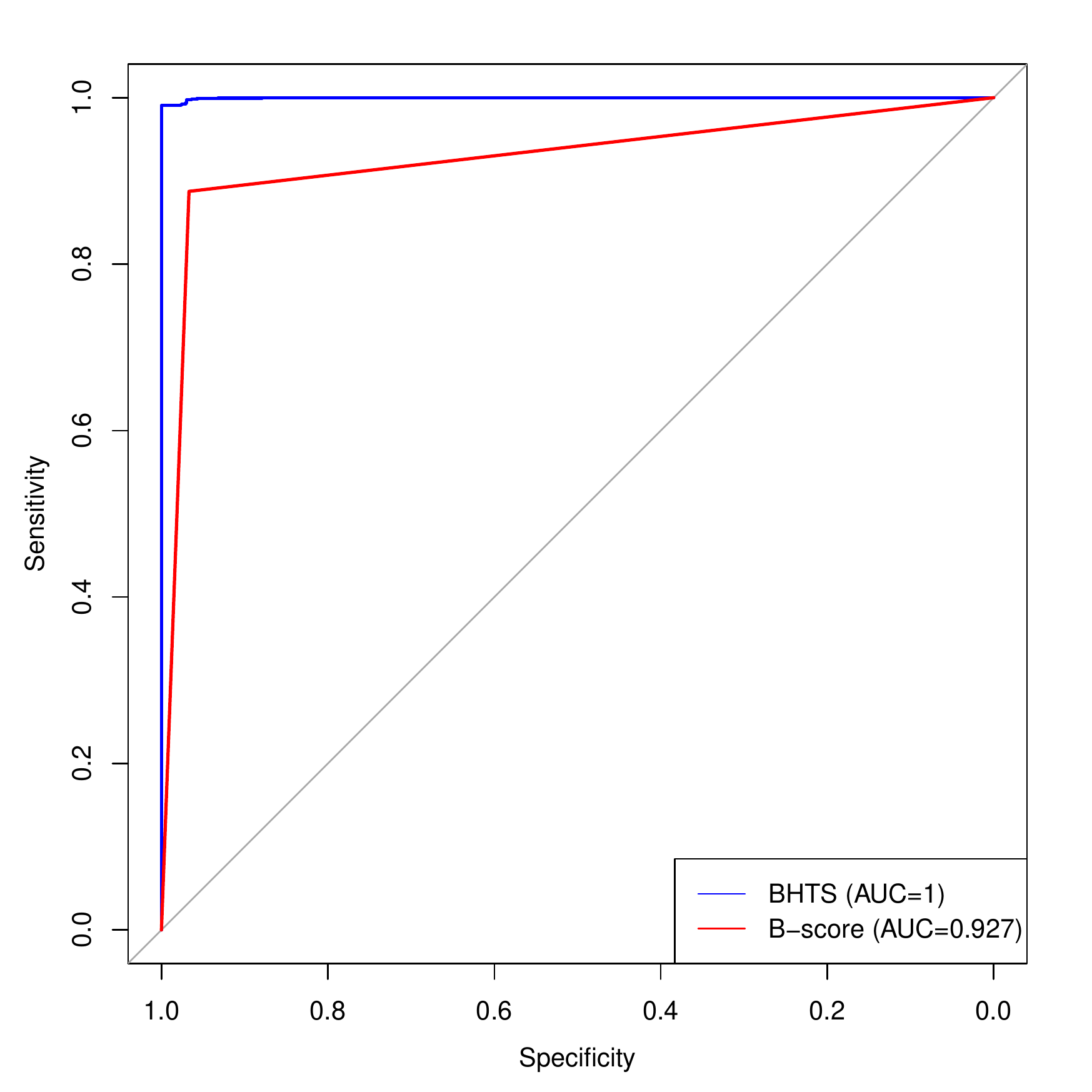}
\caption{Density of control raw values (left), B-score method AUC as a function of threshold (middle) and ROC (right) plots, based on the real control data. Data are under the auspices of NIH contract No. HHSN272201400054C.}
\label{fig:controls}
\end{figure}

We assess the proposed method capability to identify potential hits with low and medium activity by performing experiments with $688$ $96$-well plates of real must cell activated compounds and controls. Each plate contained $80$ compound and $16$ control wells. The data consisted of negative controls, low, medium and high concentration controls, positive controls, and compounds. A summary table of the data with the number of wells for each well type is shown in Tab. \ref{tab:datasummary}. The different control type densities are shown in Fig. \ref{fig:controls}. In this experiment, only control wells were used in assessing the performance of the proposed framework. The negative controls comprised the set of wells that are not hits, while the low, medium, high concentration, and positive controls comprised the set of hit wells. The rationale behind this scenario is that the data may contain compound hits with low, medium and high activity levels.

\begin{table}[h]
\centering
\scalebox{0.75}{\begin{tabular}{llrrrrr}
  \hline
Well Type & Negative & Low & Medium & High & Positive & Compounds \\ 
  \hline
Number of Wells & 3440 & 1376 & 1376 & 1376 & 3440 & 55040 \\
  \hline
\end{tabular}}
\caption{Summary of real compound and control data used in the experiments.} 
\label{tab:datasummary}
\end{table}

Experimental ROC results are shown in Fig. \ref{fig:controls}. The B-score ROC curve is based on the maximum achievable AUC threshold shown in the same figure. It can be seen that the BHTS method outperforms the B-score method in the classification of wells with low and medium concentration. This experimental comparison suggests that the BHTS approach would be more valuable in identifying compounds with low and medium activity. See supplementary for specifically chosen hyperparameters regarding this experiment.

\section{Implementation and Scalability}
We implemented the proposed model as an {\tt R} package {\tt BHTSpack} \citep{shterev2}, with some of the inner routines implemented in {\tt C/C++}. We experimented on a laptop with a Intel(R) Core(TM) i7-4600M CPU @ 2.90GHz and 8GB of RAM, running the 64-bit Ubuntu Linux operating system. It takes $30$ minutes to complete $7\times10^3$ Gibbs sampler iterations, using $10^3$ plates with $80\times10^3$ compounds. Additionally, it takes $23$ hours to complete the same number of iterations, using $50\times10^3$ plates with $4\times10^6$ compounds. The proposed MCMC algorithm for Bayesian posterior computation had good mixing rates across cases in our experiments.

\section{Conclusions}
We developed a new probabilistic framework for primary hit screening of compounds based on Bayesian statistics. The statistical model is capable of simultaneously identifying hits from multiple plates, with possibly different numbers of unique compounds, and without the use of controls. It selectively shares statistical strength among different regions of plates, thus being able to characterize systematic experimental effects across plate groups. The nonparametric nature of the model makes it suitable for handling real compound data that are not necessarily Gaussian distributed. The probabilistic hit identification rules of the algorithm facilitate principled statistical hit identification and FDR control. Experimental validation with synthetic compound data show improved sensitivity and specificity over the B-score method, which is shown to be highly sensitive to the choice of an optimal threshold. In addition, experiments with real compound data containing negative and positive controls, as well as controls with low, medium and high concentration demonstrate significant improvement of classification accuracy over the B-score method, particularly in identifying hits with low and medium activity. An efficient implementation in the form of an {\tt R} extension package {\tt BHTSpack} \citep{shterev2} makes the method applicable to large scale HTS data analysis.

\section{Acknowledgement}
This project was funded by the Division of Allergy, Immunology, and Transplantation, National Institute of Allergy and Infectious Diseases, National Institutes of Health, Department of Health and Human Services, under contract No. HHSN272201400054C entitled \textquotedblleft{Adjuvant Discovery For Vaccines Against West Nile Virus and Influenza}\textquotedblright, awarded to Duke University and lead by Drs. Herman Staats and Soman Abraham.

\bibliographystyle{plain}
\bibliography{BHTSbibl}

\end{document}

\title{Bayesian Multi Plate High Throughput Screening of Compounds: Supplementary Material}
\author[1]{Ivo D. Shterev\thanks{Corresponding Author: Ivo D. Shterev, PhD; email: {\tt i.shterev@duke.edu}}}
\author[2]{David B. Dunson}
\author[3]{Cliburn Chan}
\author[1]{Gregory D. Sempowski}
\affil[1]{Duke Human Vaccine Institute, Duke University}
\affil[2]{Department of Statistical Science, Duke University}
\affil[3]{Department of Biostatistics and Bioinformatics, Duke University}
\maketitle
\maketitle

\appendix
\section{Finite Stick-Breaking DP Representation}
This section provides a specification of the finite stick-breaking representation \citep{ishwaran} of the global $\{\boldsymbol{\lambda_K^{(1)}},\boldsymbol{\lambda_K^{(0)}}\}$ and local $\{\boldsymbol{\lambda_H^{(1)}},\boldsymbol{\lambda_H^{(0)}}\}$ DP weight priors.
\begin{eqnarray}
\boldsymbol{\lambda_K^{(1)}} & = & (\lambda_{1}^{(1)},\ldots,\lambda_{K}^{(1)}),\\
\lambda_{k}^{(1)} & = & \nu_{k}^{(1)}\prod_{l=1}^{k-1}(1-\nu_{l}^{(1)})\mbox{, for $k\in\{1,\ldots,K\}$}\\
\nu_{k}^{(1)} & \sim & \mbox{Beta}(1,\alpha_1)\mbox{, for $k\in\{1,\ldots,K\}$}
\\
\boldsymbol{\lambda_K^{(0)}} & = & (\lambda_{1}^{(0)},\ldots,\lambda_{K}^{(0)}),\\
\lambda_{k}^{(0)} & = & \nu_{k}^{(0)}\prod_{l=1}^{k-1}(1-\nu_{l}^{(0)})\mbox{, for $h\in\{1,\ldots,K\}$}\\
\nu_{k}^{(0)} & \sim & \mbox{Beta}(1,\alpha_0)\mbox{, for $h\in\{1,\ldots,K\}$}
\\
\boldsymbol{\lambda_H^{(1)}} & = & (\lambda_{1}^{(1)},\ldots,\lambda_{H}^{(1)}),\\
\lambda_{h}^{(1)} & = & \nu_{h}^{(1)}\prod_{l=1}^{h-1}(1-\nu_{l}^{(1)})\mbox{, for $h\in\{1,\ldots,H\}$}\\
\nu_{h}^{(1)} & \sim & \mbox{Beta}(1,\alpha_1)\mbox{, for $h\in\{1,\ldots,H\}$}
\\
\boldsymbol{\lambda_H^{(0)}} & = & (\lambda_{1}^{(0)},\ldots,\lambda_{H}^{(0)}),\\
\lambda_{h}^{(0)} & = & \nu_{h}^{(0)}\prod_{l=1}^{h-1}(1-\nu_{l}^{(0)})\mbox{, for $h\in\{1,\ldots,H\}$}\\
\nu_{h}^{(0)} & \sim & \mbox{Beta}(1,\alpha_0)\mbox{, for $h\in\{1,\ldots,H\}$}.
\end{eqnarray}

\clearpage
\section{MCMC Update Equations}
This section presents the MCMC update equations based on approximate full conditional posteriors of the model parameters and construct a Gibbs sampler that iteratively samples from these update equations.

Let indicator variable $I_{mi}^{(1)}\in\{1,\ldots,K\}$ specify the active component to which the $i$th compound from the $m$th plate belongs. Analogously, $I_{mi}^{(0)}\in\{1,\ldots,K\}$ specifies the inactive component to which the $i$th compound from the $m$th plate belongs.
\begin{itemize}
\item updating compound hit indicator variable $b_{mi}$.
\begin{eqnarray}
p(b_{mi}|\hat\pi(z_{mi})) & \propto & \mbox{Bernoulli}(b_{mi}|\hat\pi(z_{mi})),
\end{eqnarray}
where
\begin{eqnarray}
\hat\pi(z_{mi}) & = & \frac{1}{1+\frac{1-\pi}{\pi}\frac{\sum_h\lambda_{mh}^{(0)}\mathcal{K}\big(z_{mi};\theta_{h}^{(0)}\big)}{\sum_h\lambda_{mh}^{(1)}\mathcal{K}\big(z_{mi};\theta_{h}^{(1)}\big)}}.
\end{eqnarray}

\item updating $\pi$.
\begin{eqnarray}
p(\pi|\mbox{all }b_{mi}) & = & \mbox{Beta}(\pi|\hat{a}_\pi,\hat{b}_\pi),
\end{eqnarray}
where 
\begin{eqnarray}
\hat{a}_\pi & = & a_\pi+\sum_{m,i}b_{mi},\\
\hat{b}_\pi & = & b_\pi+\sum_{m,i}(1-b_{mi}).
\end{eqnarray}

\item updating $\theta_{k}^{(1)}=\{\mu_{1k},\sigma_{1k}^2\}$
\begin{eqnarray}
p(\mu_{1k},\sigma_{1k}^2|z_{mi}) & = & \mathcal{N}\Big(\mu_{1k}\Big|\frac{\mu_{10}+n_k^{(1)}\bar z_k^{(1)}}{n_k^{(1)}+1},\frac{\sigma_{1k}^2}{n_k^{(1)}+1}\Big)\mbox{Inv-Ga}\Big(\sigma_{1k}^2\Big|\hat{a}_{1k}, \hat{b}_{1k}\Big),\\
\end{eqnarray}
where
\begin{eqnarray}
\hat{a}_{1k} & = & a+n_k^{(1)}\\
\hat{b}_{1k} & = & b+0.5\sum_{m,i}(z_{mi}-\bar z_k^{(1)})^2+0.5\frac{n_k^{(1)}}{n_k^{(1)}+1}(\mu_{10}-\bar z_k^{(1)})^2\\
\bar z_k^{(1)} & = & \frac{1}{n_k^{(1)}}\sum_{m,i}b_{mi}1(I_{mi}^{(1)}=k)z_{mi}\\
n_k^{(1)} & = & \sum_{m,i}b_{mi}1(I_{mi}^{(1)}=k)\\
1(I_{mi}^{(1)}=k) & = &  \left\{
\begin{array}{rl}
  1 & \mbox{if } z_{mi}\mbox{ belongs to the $k$th active component},\\
  0 & \mbox{if } \mbox{otherwise}.
\end{array} \right.
\end{eqnarray}

\item updating $\theta_{k}^{(0)}=\{\mu_{0k},\sigma_{0k}^2\}$
\begin{eqnarray}
p(\mu_{0k},\sigma_{0k}^2|z_{mi}) & = & \mathcal{N}\Big(\mu_{0k}\Big|\frac{\mu_{10}+n_k^{(0)}\bar z_k^{(0)}}{n_k^{(0)}+1},\frac{\sigma_{0k}^2}{n_k^{(0)}+1}\Big)\mbox{Inv-Ga}\Big(\sigma_{0k}^2\Big|\hat{a}_{0k}, \hat{b}_{0k}\Big),\\
\end{eqnarray}
where
\begin{eqnarray}
\hat{a}_{0k} & = & a+n_k^{(0)}\\
\hat{b}_{0k} & = & b+0.5\sum_{m,i}(z_{mi}-\bar z_k^{(0)})^2+0.5\frac{n_k^{(0)}}{n_k^{(0)}+1}(\mu_{00}-\bar z_k^{(0)})^2\\
\bar z_k^{(0)} & = & \frac{1}{n_k^{(0)}}\sum_{m,i}(1-b_{mi})1(I_{mi}^{(0)}=k)z_{mi}\\
n_k^{(0)} & = & \sum_{m,i}(1-b_{mi})1(I_{mi}^{(0)}=k)\\
1(I_{mi}^{(0)}=k) & = &  \left\{
\begin{array}{rl}
  1 & \mbox{if } z_{mi}\mbox{ belongs to the $k$th inactive component},\\
  0 & \mbox{if } \mbox{otherwise}.
\end{array} \right.
\end{eqnarray}

\item updating $\lambda_{mh}^{(1)}$.
\begin{eqnarray}
\lambda_{mh}^{(1)} & = & \nu_{mh}^{(1)}\prod_{l=1}^{h-1}(1-\nu_{ml}^{(1)})\\
\nu_{mh}^{(1)} & \sim & \mbox{Beta}(1+p_{mh}^{(1)}, \alpha_1+\sum_{l=h+1}^Hp_{ml}^{(1)})\\
p_{mh}^{(1)} & = & \sum_ib_{mi}1(J_{mi}^{(1)}=h)\\
1(J_{mi}^{(1)}=h) & = &  \left\{
\begin{array}{rl}
  1 & \mbox{if } z_{mi}\mbox{ belongs to the $h$th active cluster},\\
  0 & \mbox{if } \mbox{otherwise}.
\end{array} \right.
\end{eqnarray}

\item updating $\lambda_{mh}^{(0)}$.
\begin{eqnarray}
\lambda_{mh}^{(0)} & = & \nu_{mh}^{(0)}\prod_{l=1}^{h-1}(1-\nu_{ml}^{(0)})\\
\nu_{mh}^{(0)} & \sim & \mbox{Beta}(1+p_{mh}^{(0)}, \alpha_0+\sum_{l=h+1}^Hp_{ml}^{(0)})\\
p_{mh}^{(0)} & = & \sum_i(1-b_{mi})1(J_{mi}^{(0)}=h)\\
1(J_{mi}^{(0)}=h) & = &  \left\{
\begin{array}{rl}
  1 & \mbox{if } z_{mi}\mbox{ belongs to the $h$th inactive cluster},\\
  0 & \mbox{if } \mbox{otherwise}.
\end{array} \right.
\end{eqnarray}

\item updating $\lambda_{k}^{(1)}$.
\begin{eqnarray}
\lambda_{k}^{(1)} & = & \nu_{k}^{(1)}\prod_{l=1}^{k-1}(1-\nu_{l}^{(1)})\\
\nu_{k}^{(1)} & \sim & \mbox{Beta}(1+p_{k}^{(1)}, \tau_1+\sum_{l=k+1}^Kp_{l}^{(1)})\\
p_{k}^{(1)} & = & \sum_{m,h}1(J_{mh}^{(1)}=k)\\
1(J_{mh}^{(1)}=k) & = &  \left\{
\begin{array}{rl}
  1 & \mbox{if } \mbox{cluster $h$ belongs to the $k$th active component},\\
  0 & \mbox{if } \mbox{otherwise}.
\end{array} \right.
\end{eqnarray}

\item updating $\lambda_{k}^{(0)}$.
\begin{eqnarray}
\lambda_{k}^{(0)} & = & \nu_{k}^{(0)}\prod_{l=1}^{k-1}(1-\nu_{l}^{(0)})\\
\nu_{k}^{(0)} & \sim & \mbox{Beta}(1+p_{k}^{(0)}, \tau_0+\sum_{l=k+1}^Kp_{l}^{(0)})\\
p_{k}^{(0)} & = & \sum_{m,h}1(J_{mh}^{(0)}=k)\\
1(J_{mh}^{(0)}=k) & = &  \left\{
\begin{array}{rl}
  1 & \mbox{if } \mbox{cluster $h$ belongs to the $k$th inactive component},\\
  0 & \mbox{if } \mbox{otherwise}.
\end{array} \right.
\end{eqnarray}

\item updating $\alpha_1$.
\begin{eqnarray}
p(\alpha_1|\mbox{all } \nu_{mh}^{(1)}) & \propto & \mbox{Ga}\Big(\alpha_1|a_{\alpha}+M(H-1), b_{\alpha}-\sum_m\sum_{h=1}^{H-1}\ln(1-\nu_{mh}^{(1)})\Big)
\end{eqnarray}

\item updating $\alpha_0$.
\begin{eqnarray}
p(\alpha_0|\mbox{all } \nu_{mh}^{(0)}) & \propto & \mbox{Ga}\Big(\alpha_0|a_{\alpha}+M(H-1), b_{\alpha}-\sum_m\sum_{h=1}^{H-1}\ln(1-\nu_{mh}^{(0)})\Big)
\end{eqnarray}

\item updating $\tau_1$.
\begin{eqnarray}
p(\tau_1|\mbox{all } \nu_{k}^{(1)}) & \propto & \mbox{Ga}\Big(\tau_1|a_{\tau}+K-1, b_{\tau}-\sum_{k=1}^{K-1}\ln(1-\nu_{k}^{(1)})\Big)
\end{eqnarray}

\item updating $\tau_0$.
\begin{eqnarray}
p(\tau_0|\mbox{all } \nu_{k}^{(0)}) & \propto & \mbox{Ga}\Big(\tau_0|a_{\tau}+K-1, b_{\tau}-\sum_{k=1}^{K-1}\ln(1-\nu_{k}^{(0)})\Big)
\end{eqnarray}

\item updating $J_{mi}^{(1)}$.
\begin{eqnarray}
J_{mi}^{(1)} & \sim & \mbox{Mult}(\lambda_{m1}^{(1)}\mathcal{K}\big(z_{mi};\theta_{J_{m1}}^{(1)}\big),\ldots,\lambda_{mH}^{(1)}\mathcal{K}\big(z_{mi};\theta_{J_{mH}}^{(1)}\big))
\end{eqnarray}

\item updating $J_{mi}^{(0)}$.
\begin{eqnarray}
J_{mi}^{(0)} & \sim & \mbox{Mult}(\lambda_{m1}^{(0)}\mathcal{K}\big(z_{mi};\theta_{J_{m1}}^{(0)}\big),\ldots,\lambda_{mH}^{(0)}\mathcal{K}\big(z_{mi};\theta_{J_{mH}}^{(0)}\big))
\end{eqnarray}

\item updating $J_{mh}^{(1)}$.
\begin{eqnarray}
J_{mh}^{(1)} & \sim & \mbox{Mult}(\lambda_{k}^{(1)}\prod_{i}\mathcal{K}\big(z_{mi};\theta_{1}^{(1)}\big)^{1(J_{mi}=h)},\ldots,\lambda_{K}^{(1)}\prod_{i}\mathcal{K}\big(z_{mi};\theta_{K}^{(1)}\big)^{1(J_{mi}=h)})
\end{eqnarray}

\item updating $J_{mh}^{(0)}$.
\begin{eqnarray}
J_{mh}^{(0)} & \sim & \mbox{Mult}(\lambda_{k}^{(0)}\prod_{i}\mathcal{K}\big(z_{mi};\theta_{1}^{(0)}\big)^{1(J_{mi}=h)},\ldots,\lambda_{K}^{(0)}\prod_{i}\mathcal{K}\big(z_{mi};\theta_{K}^{(0)}\big)^{1(J_{mi}=h)}),
\end{eqnarray}
\end{itemize}
where $\mbox{Mult}(\cdot)$ denotes the multinomial distribution. We have relaxed the notation by allowing the indicator variable $J$ to specify both data points (compounds) and cluster indicators, depending on the subscript\footnote{Subscript $mi$ spans over plates and compounds (hence $J_{mi}$ is a compound indicator), while subscript $mh$ spans over plates and clusters (hence $J_{mh}$ is a cluster indicator).}. We also drop the superscript of $J$ where it is clear from the context.

\clearpage
\section{Choice of Hyperparameters}
This section describes the choice of hyperparameters, for the synthetic and real control data set experiments. In all experiments, gamma density hyperparameters were fixed as $a_{\alpha_0}=a_{\alpha_1}=a_{\tau_0}=a_{\tau_1}=10$ and $b_{\alpha_0}=b_{\alpha_1}=b_{\tau_0}=b_{\tau_1}=5$. The active compound proportion prior hyperparameters were fixed as $a_{\pi}=b_{\pi}=0.5\sum_mn_m$. The hyperparameters specifying the number of plate specific clusters and global components were fixed as $H=10$ and $K=10$, respectively. The hyperparameters $a$ and $b$ can be computed based on the variance of the data $v$, as described in the paper via equations (14) and (15). The hyperparameters $\mu_{10}$, $\mu_{00}$, $\{a,b\}$ used in the experiments are summarized in Table \ref{tab:paramsummary}.

\begin{table}[h]
\centering
\scalebox{0.75}{\begin{tabular}{llrrrr}
  \hline
Experiment & $v$ & $a$ & $b$ & $\mu_{10}$ & $\mu_{00}$ \\ 
  \hline
Synthetic (40\% active compounds) & 0.01573718 & 4.476587 & 0.05471166 & 0.26 & 0.13 \\
  \hline
Synthetic (10\% active compounds) & 0.01544984 & 4.386977 & 0.05232827 & 0.26 & 0.13 \\
  \hline
Synthetic (5\% active compounds) & 0.01508506 & 4.27559 & 0.04941247 & 0.26 & 0.13 \\
  \hline
Real Compounds and Controls & 0.4031341 & 1627.171 & 655.5651 & 1.890824 & 0.09987064 \\
  \hline
\end{tabular}}
\caption{Summary of data statistics and hyperparameters used in the experiments.}
\label{tab:paramsummary}
\end{table}

\clearpage
\section{Model Convergence and Mixing}
To assess convergence and mixing of the Gibbs sampler we provide trace plots of some latent variables. We performed a total of $7000$ iterations, discarding the first $3500$ samples and using the remaining $3500$ samples. We show trace plots of the global parameters $\mu_{1k}$, $\mu_{0k}$, $\sigma_{1k}^2$, $\sigma_{0k}^2$, and the global mixing components $\lambda_{k}^{(1)}$ and $\lambda_{k}^{(0)}$, for each of the synthetically generated data sets. 

\clearpage
\subsection{Synthetic Data Containing $40\%$ Active Compounds}

\begin{figure}[h]
\centering
\includegraphics[scale=0.35]{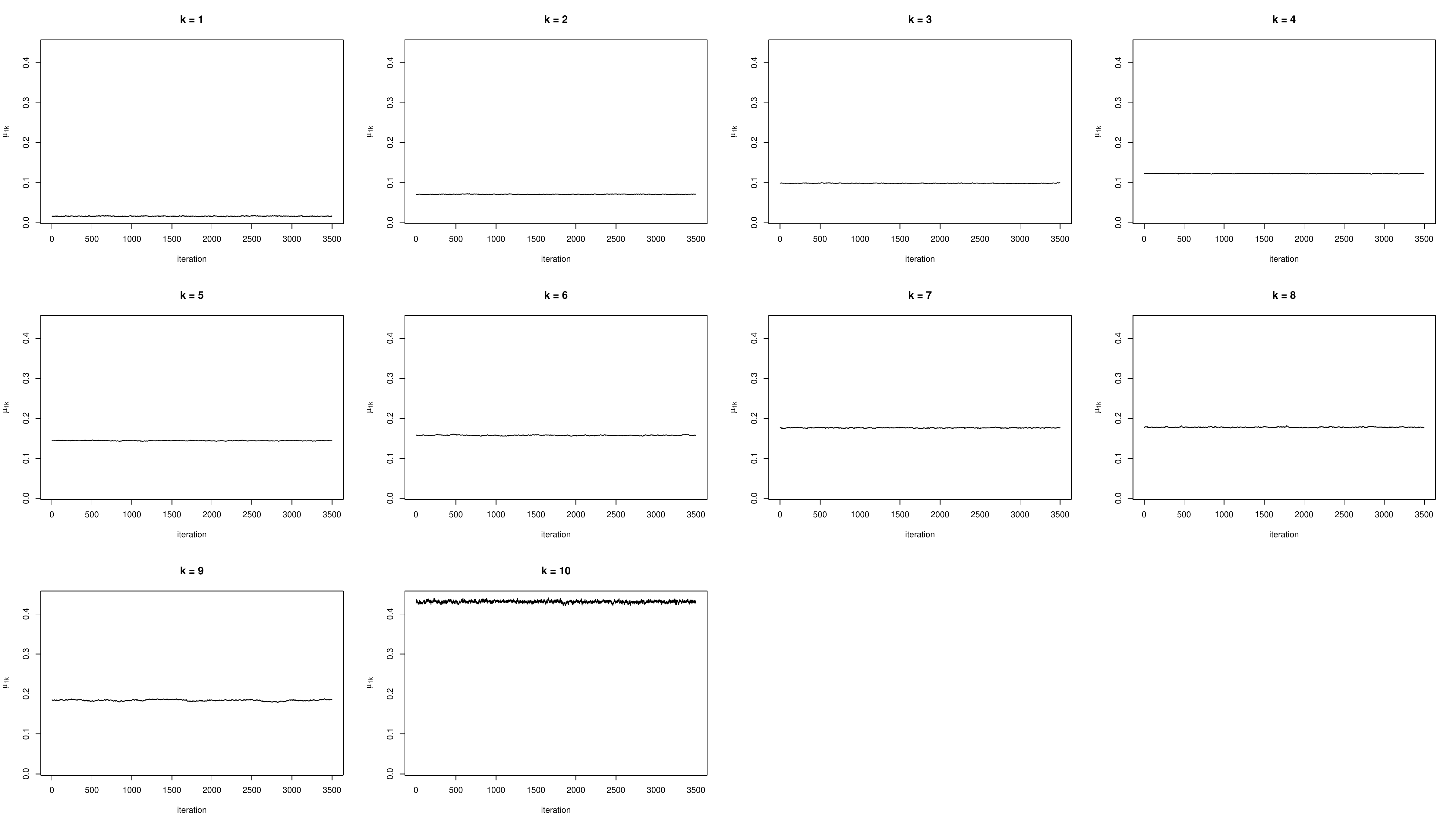}
\caption{Trace plots of $\mu_{1k}$. At each iteration, the means were sorted in increasing order to avoid label switching.}
\label{fig:mu1_traceplot}
\end{figure} 

\begin{figure}[h]
\centering
\includegraphics[scale=0.35]{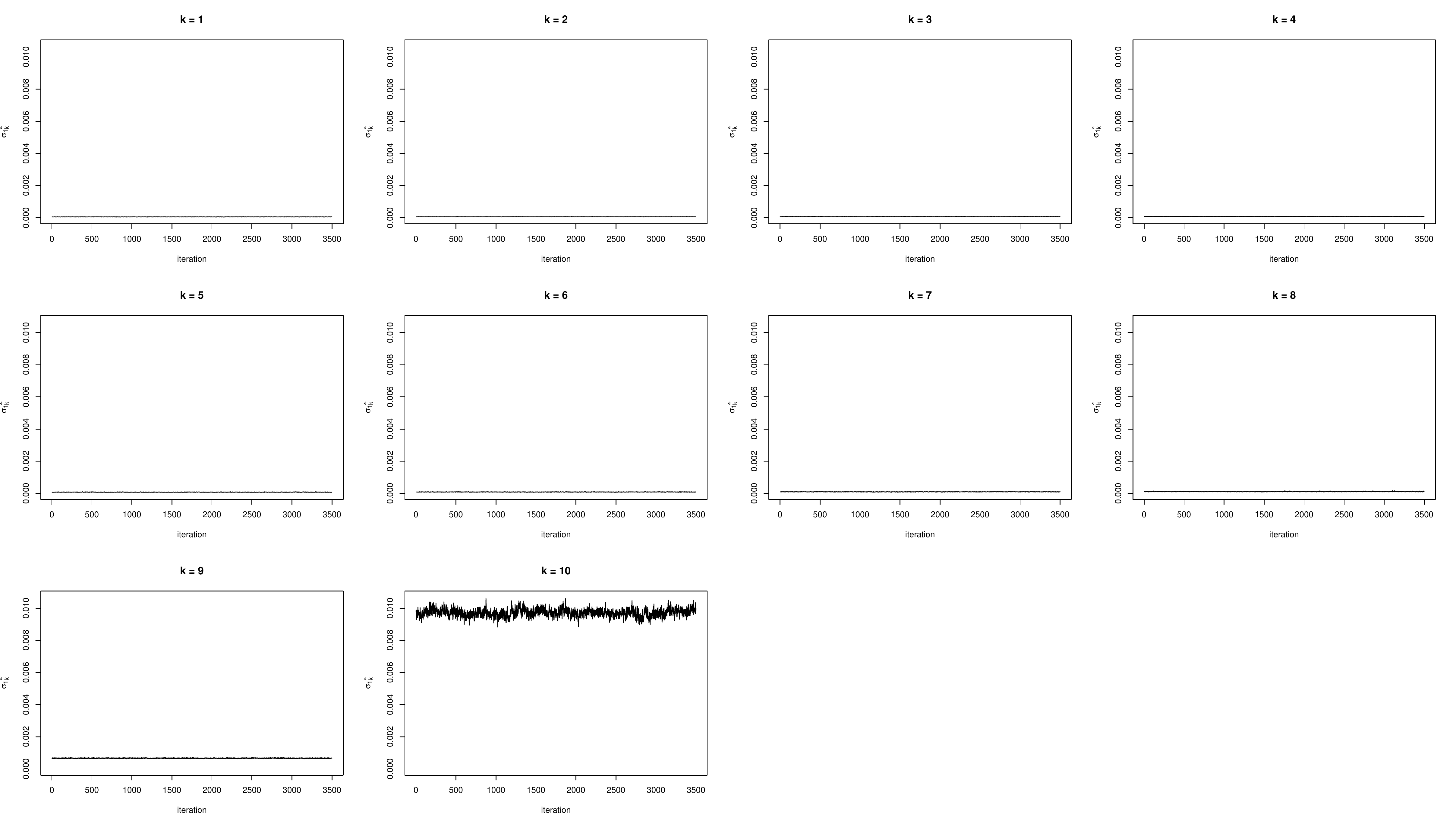}
\caption{Trace plots of $\sigma_{1k}^2$. At each iteration, the variances were sorted in increasing order to avoid label switching.}
\label{fig:sigma1_traceplot}
\end{figure} 

\begin{figure}[h]
\centering
\includegraphics[scale=0.35]{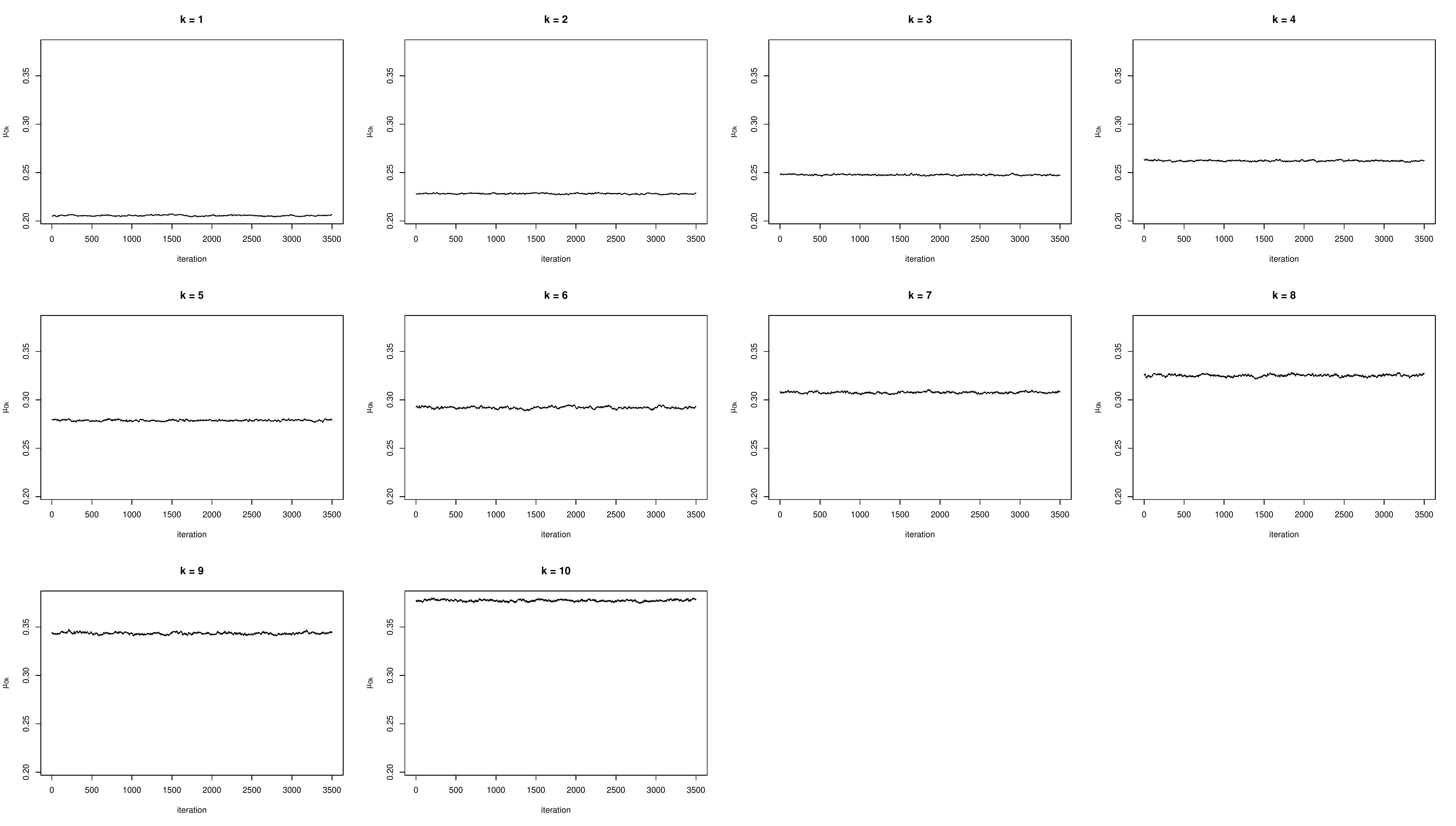}
\caption{Trace plots of $\mu_{0k}$. At each iteration, the means were sorted in increasing order to avoid label switching.}
\label{fig:mu0_traceplot}
\end{figure} 

\begin{figure}[h]
\centering
\includegraphics[scale=0.35]{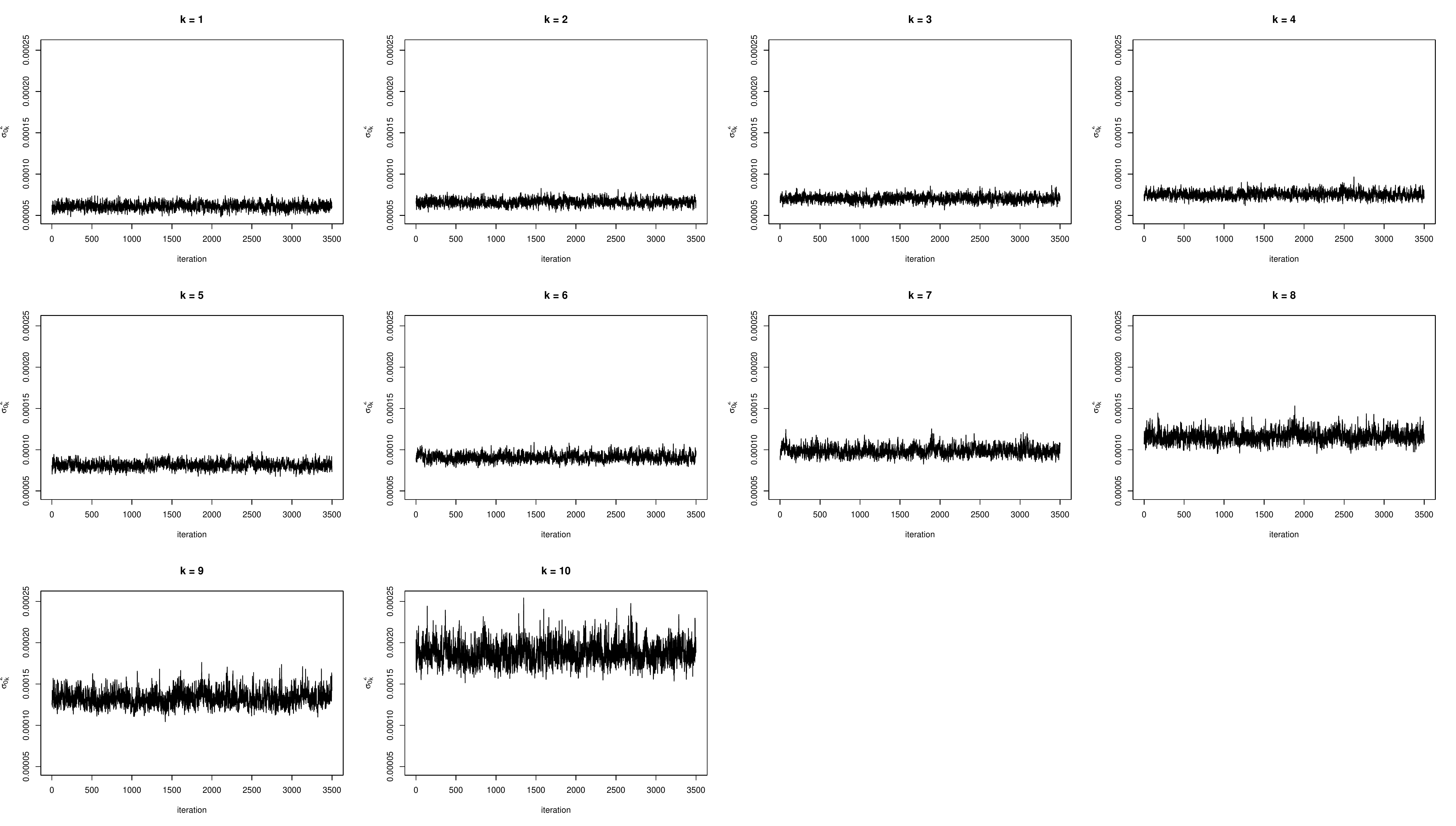}
\caption{Trace plots of $\sigma_{0k}^2$. At each iteration, the variances were sorted in increasing order to avoid label switching.}
\label{fig:sigma0_traceplot}
\end{figure} 

\begin{figure}[h]
\centering
\includegraphics[scale=0.35]{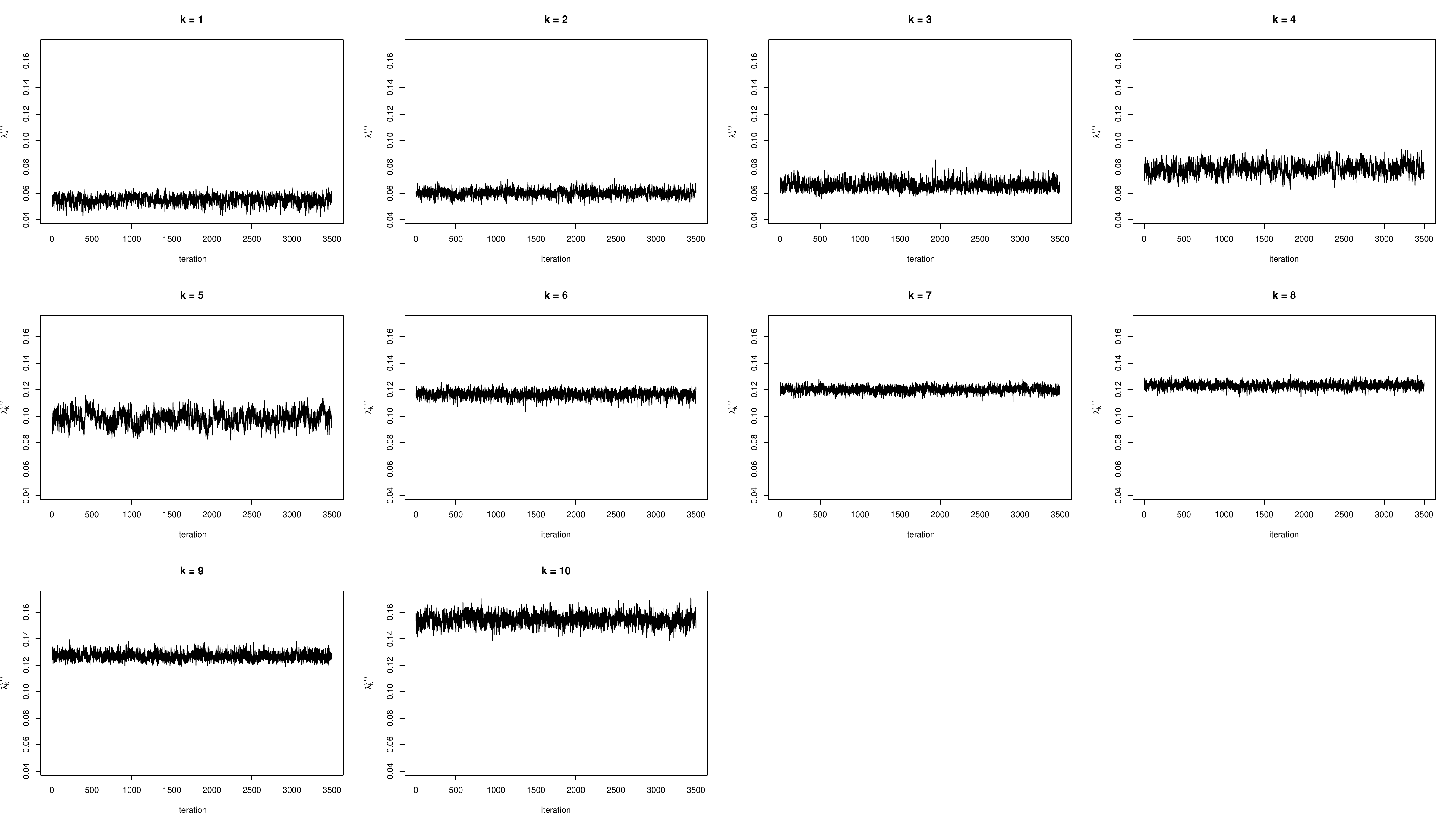}
\caption{Trace plots of $\lambda_{k}^{(1)}$.}
\label{fig:pk1_traceplot}
\end{figure} 

\begin{figure}[h]
\centering
\includegraphics[scale=0.35]{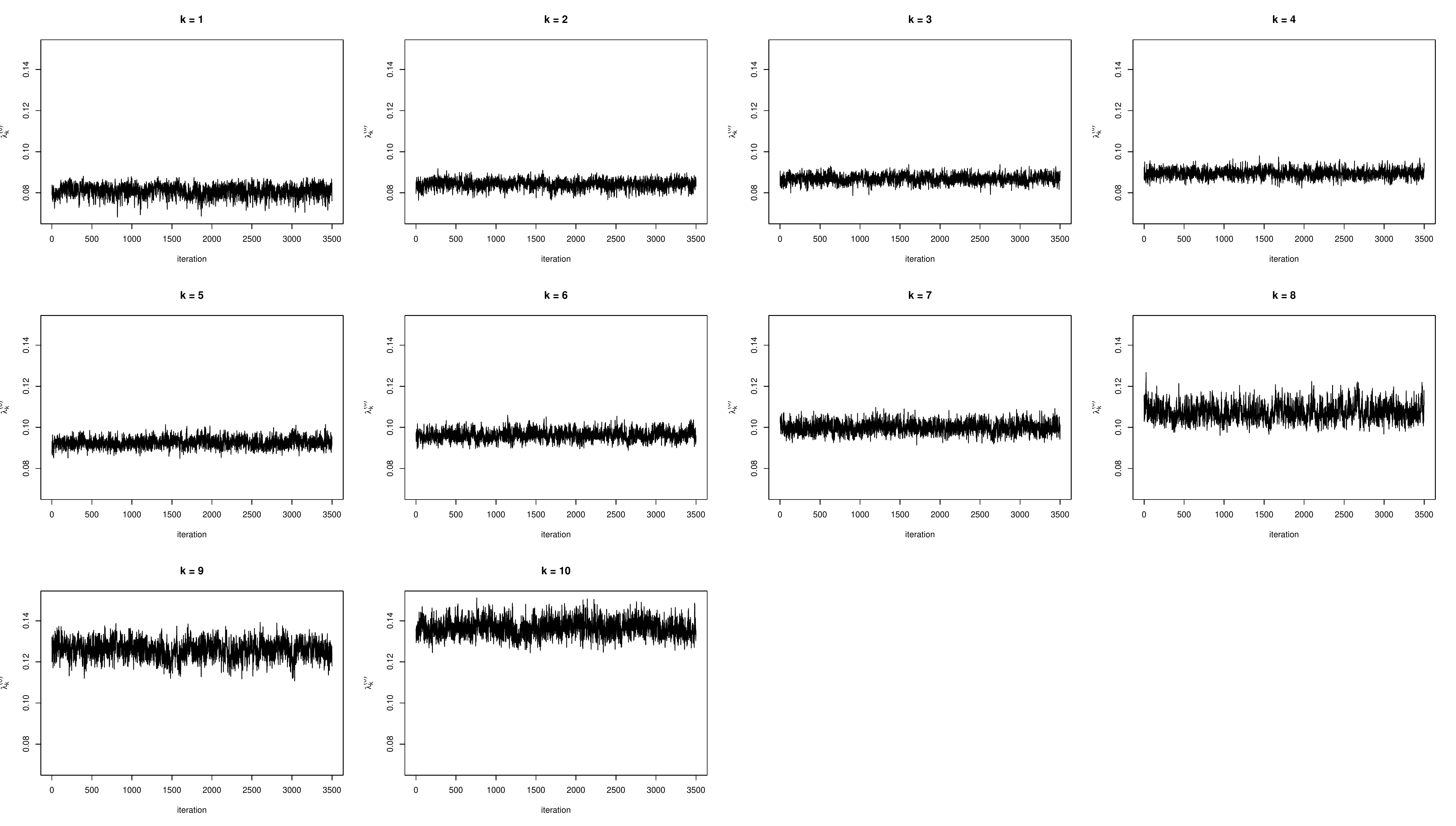}
\caption{Trace plots of $\lambda_{k}^{(0)}$.}
\label{fig:pk0_traceplot}
\end{figure}

\clearpage
\subsection{Synthetic Data Containing $10\%$ Active Compounds}

\begin{figure}[h]
\centering
\includegraphics[scale=0.35]{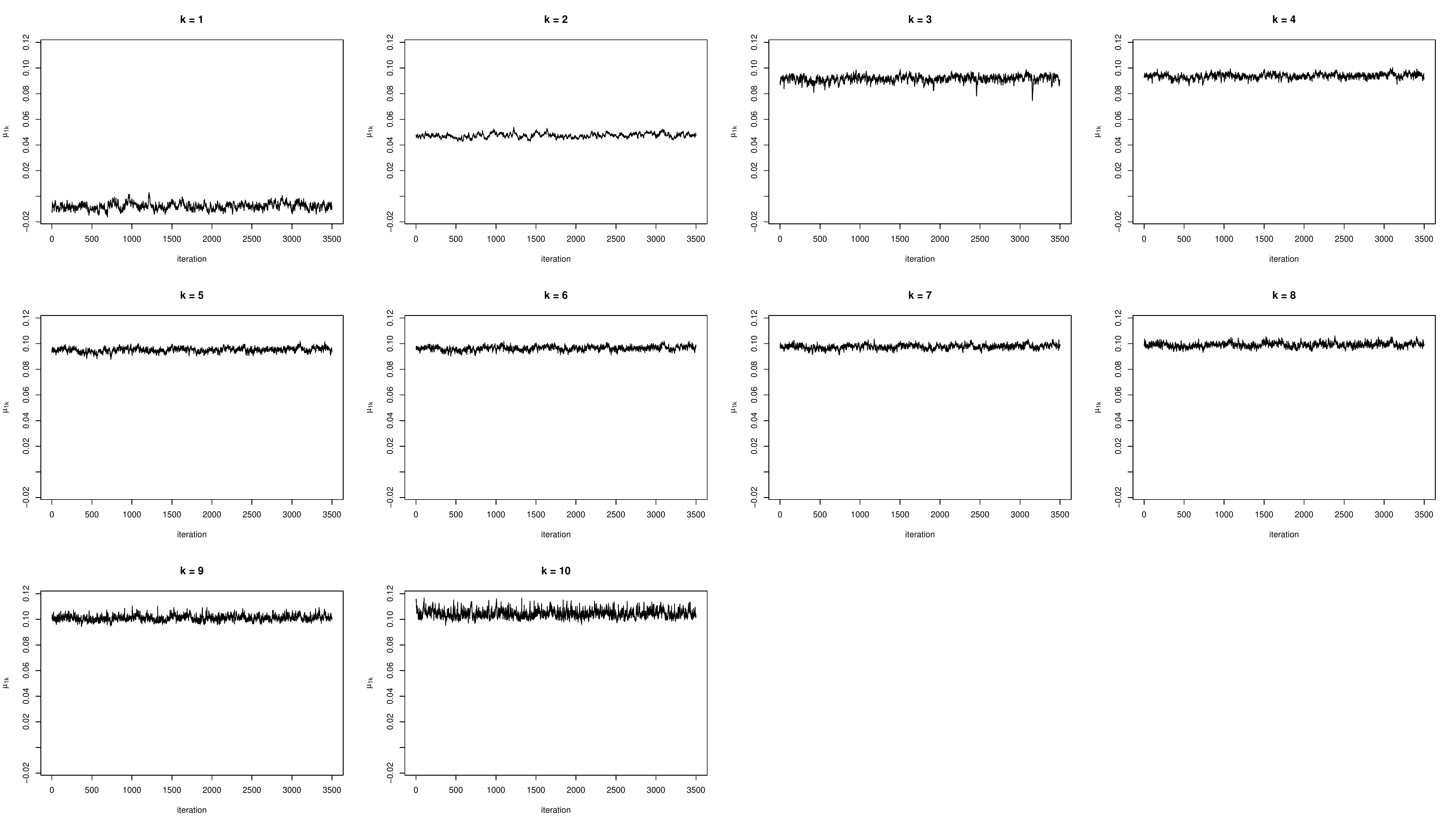}
\caption{Trace plots of $\mu_{1k}$. At each iteration, the means were sorted in increasing order to avoid label switching.}
\label{fig:mu1_traceplot}
\end{figure} 

\begin{figure}[h]
\centering
\includegraphics[scale=0.35]{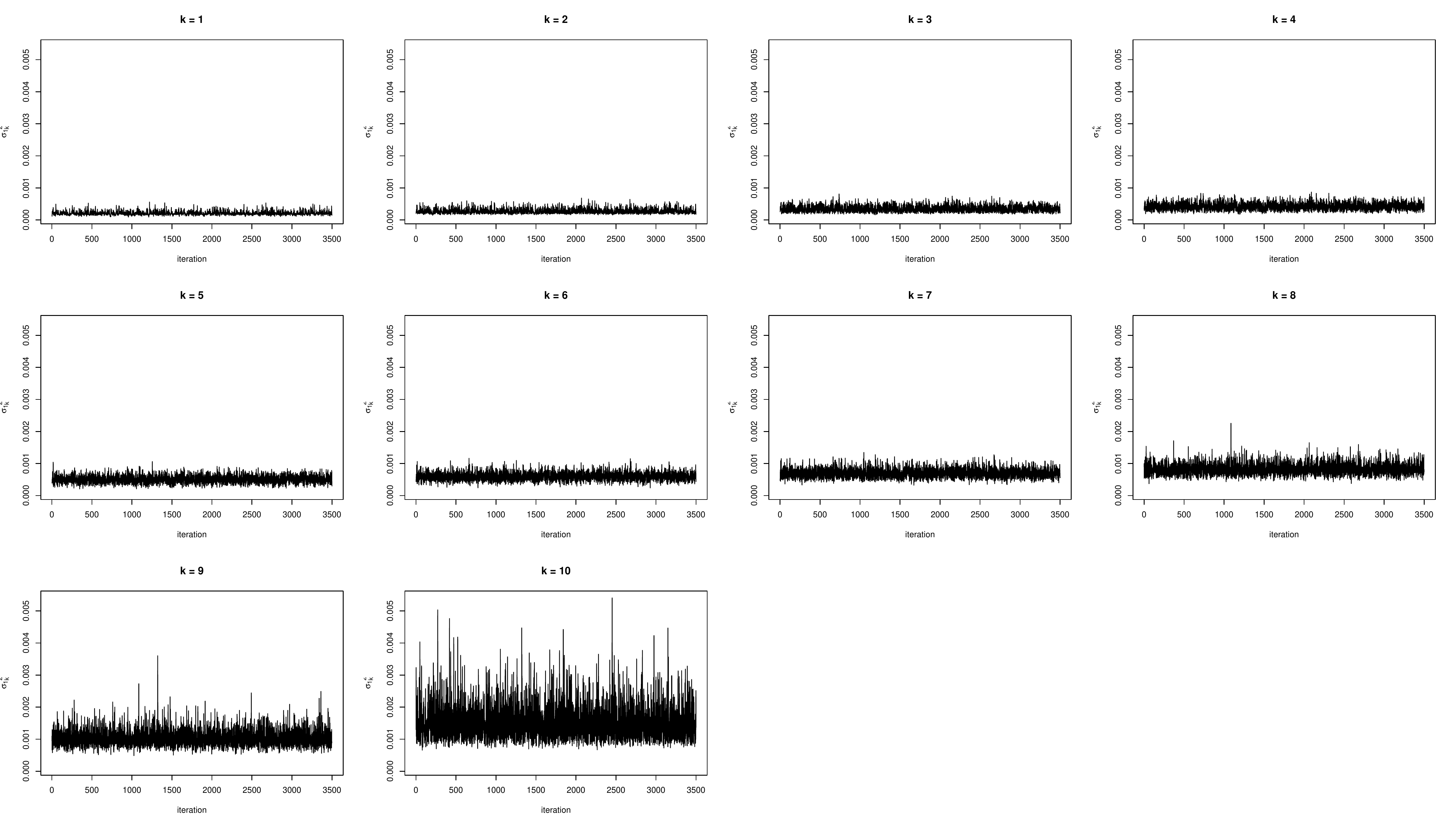}
\caption{Trace plots of $\sigma_{1k}^2$. At each iteration, the variances were sorted in increasing order to avoid label switching.}
\label{fig:sigma1_traceplot}
\end{figure} 

\begin{figure}[h]
\centering
\includegraphics[scale=0.35]{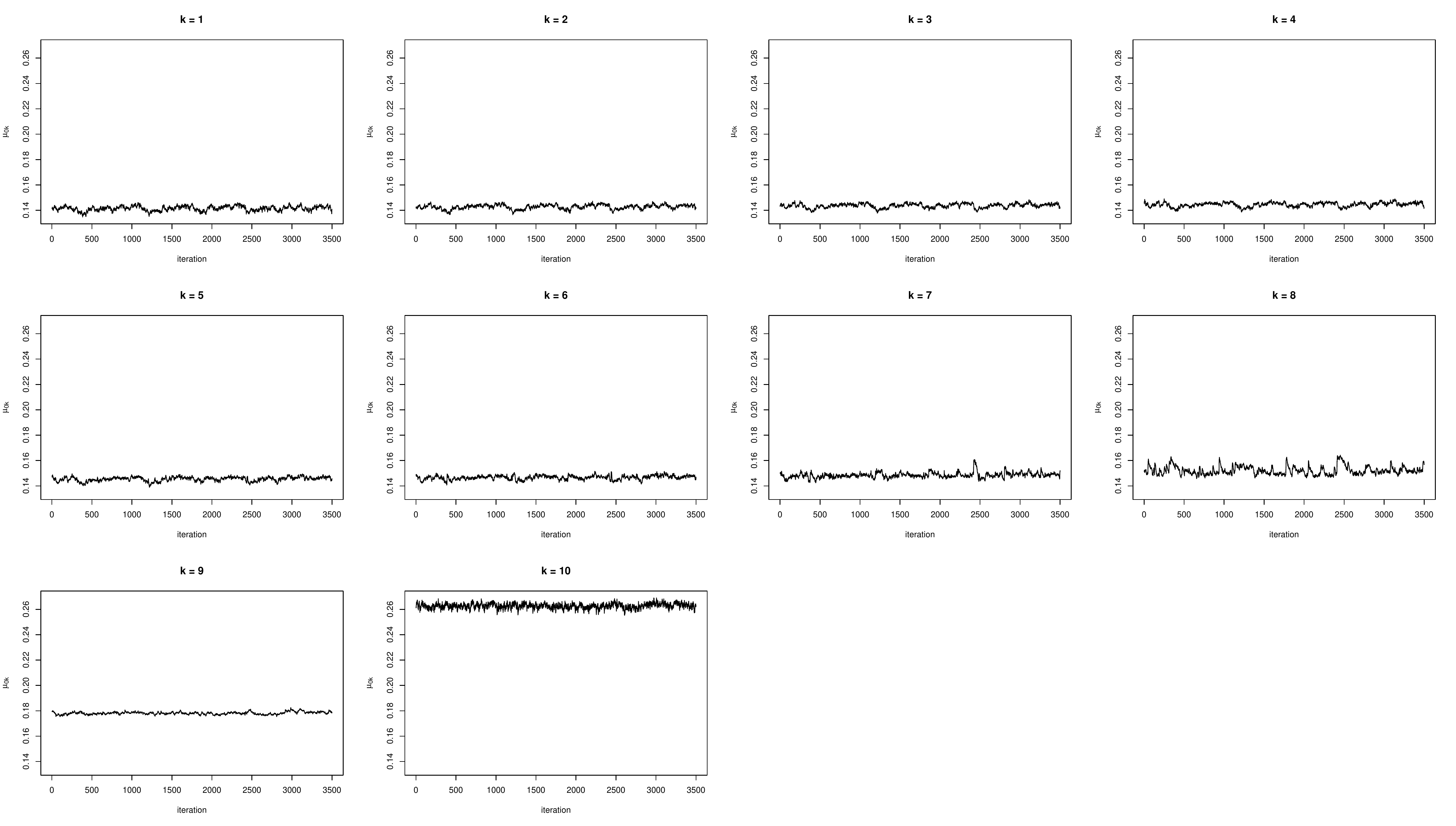}
\caption{Trace plots of $\mu_{0k}$. At each iteration, the means were sorted in increasing order to avoid label switching.}
\label{fig:mu0_traceplot}
\end{figure} 

\begin{figure}[h]
\centering
\includegraphics[scale=0.35]{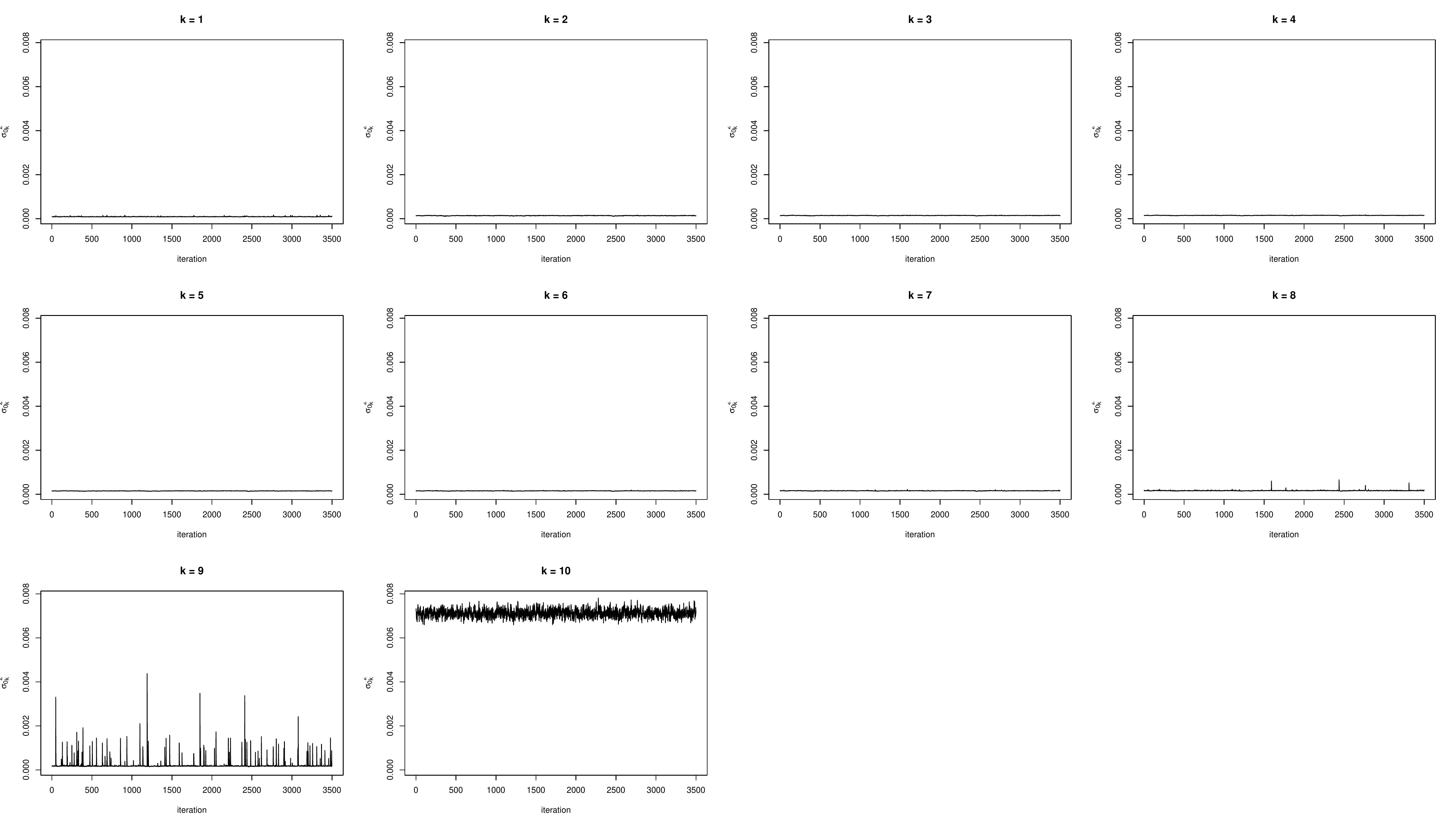}
\caption{Trace plots of $\sigma_{0k}^2$. At each iteration, the variances were sorted in increasing order to avoid label switching.}
\label{fig:sigma0_traceplot}
\end{figure} 

\begin{figure}[h]
\centering
\includegraphics[scale=0.35]{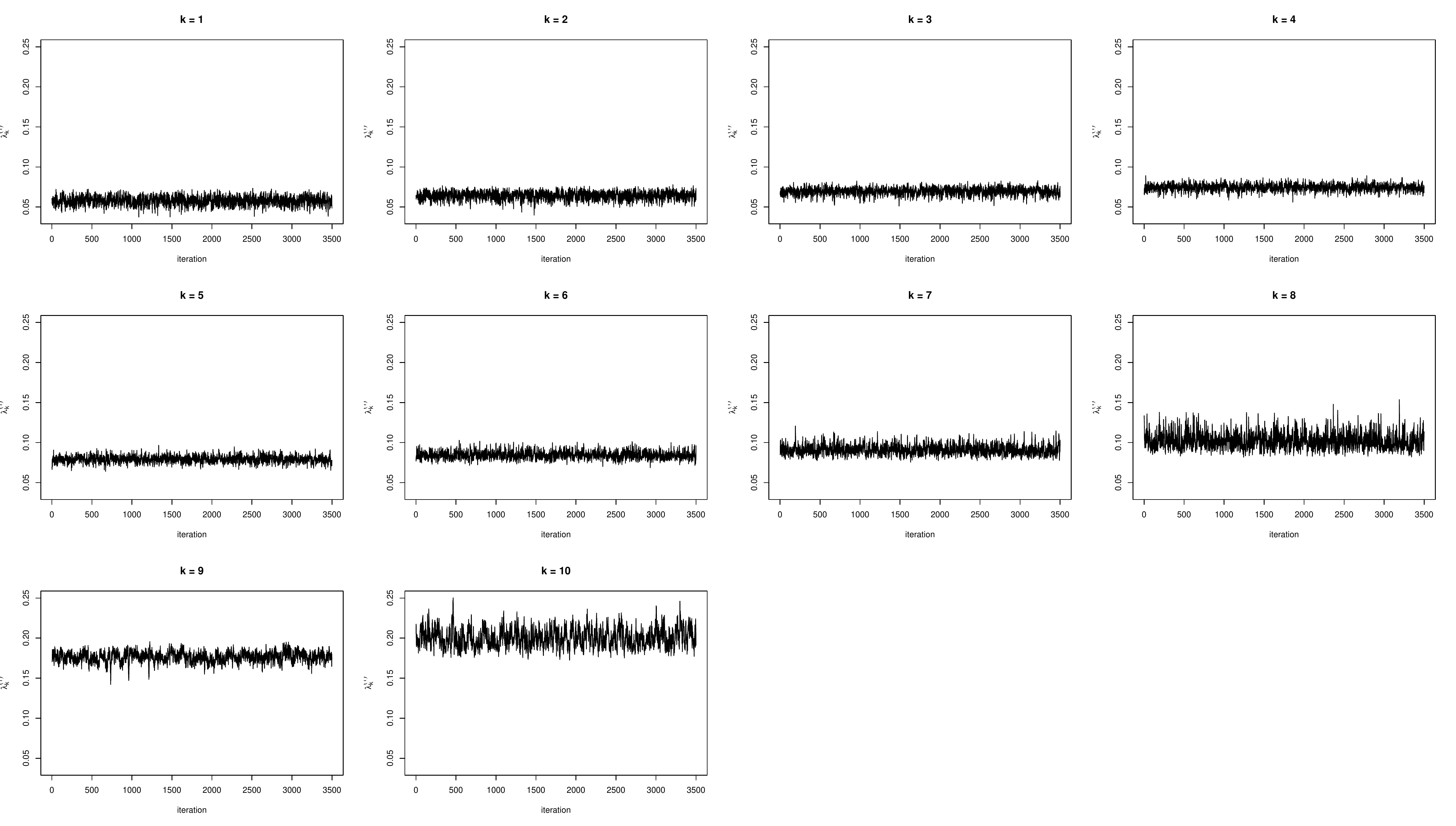}
\caption{Trace plots of $\lambda_{k}^{(1)}$.}
\label{fig:pk1_traceplot}
\end{figure} 

\begin{figure}[h]
\centering
\includegraphics[scale=0.35]{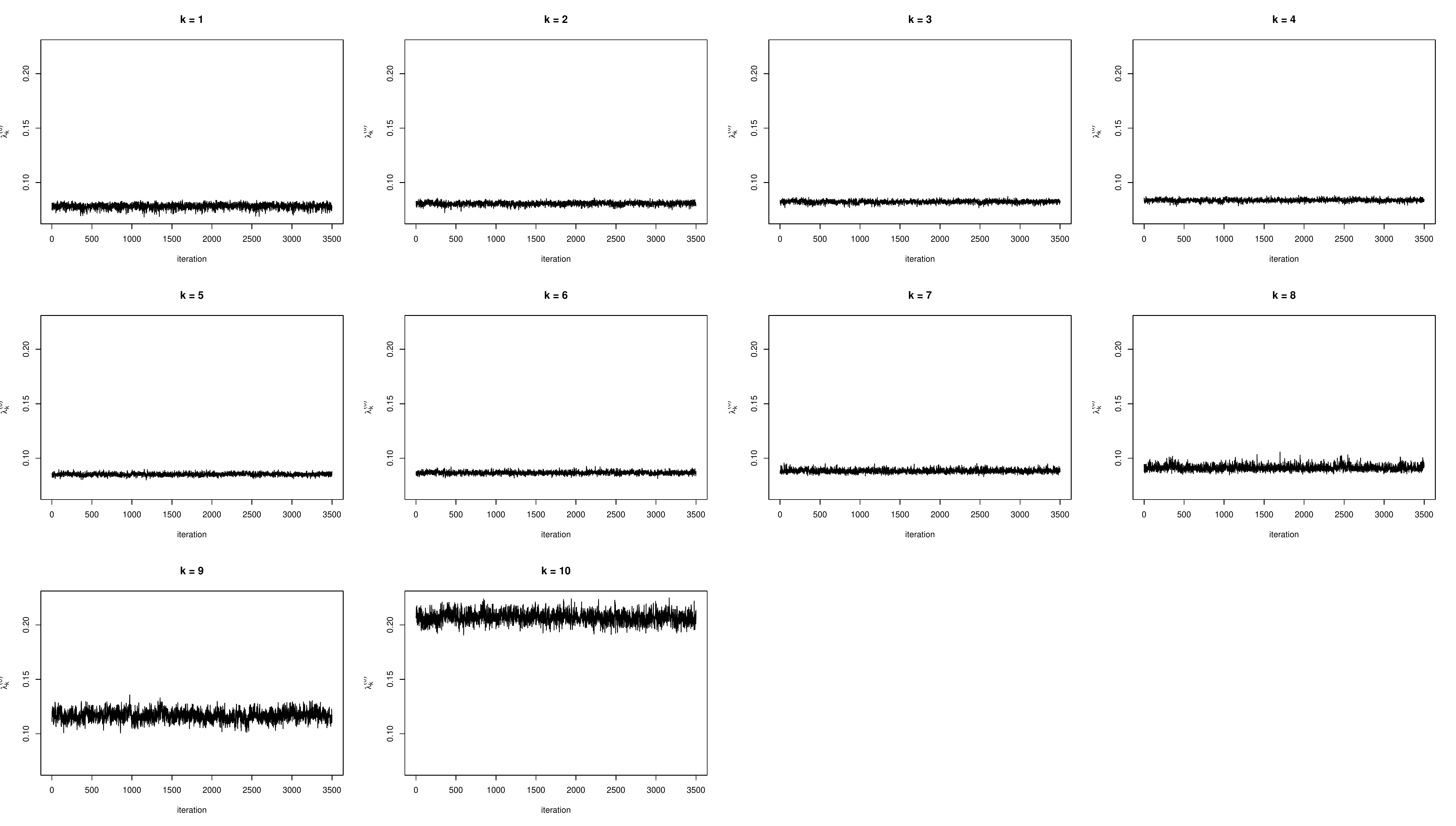}
\caption{Trace plots of $\lambda_{k}^{(0)}$.}
\label{fig:pk0_traceplot}
\end{figure}

\clearpage
\subsection{Synthetic Data Containing $5\%$ Active Compounds}

\begin{figure}[h]
\centering
\includegraphics[scale=0.35]{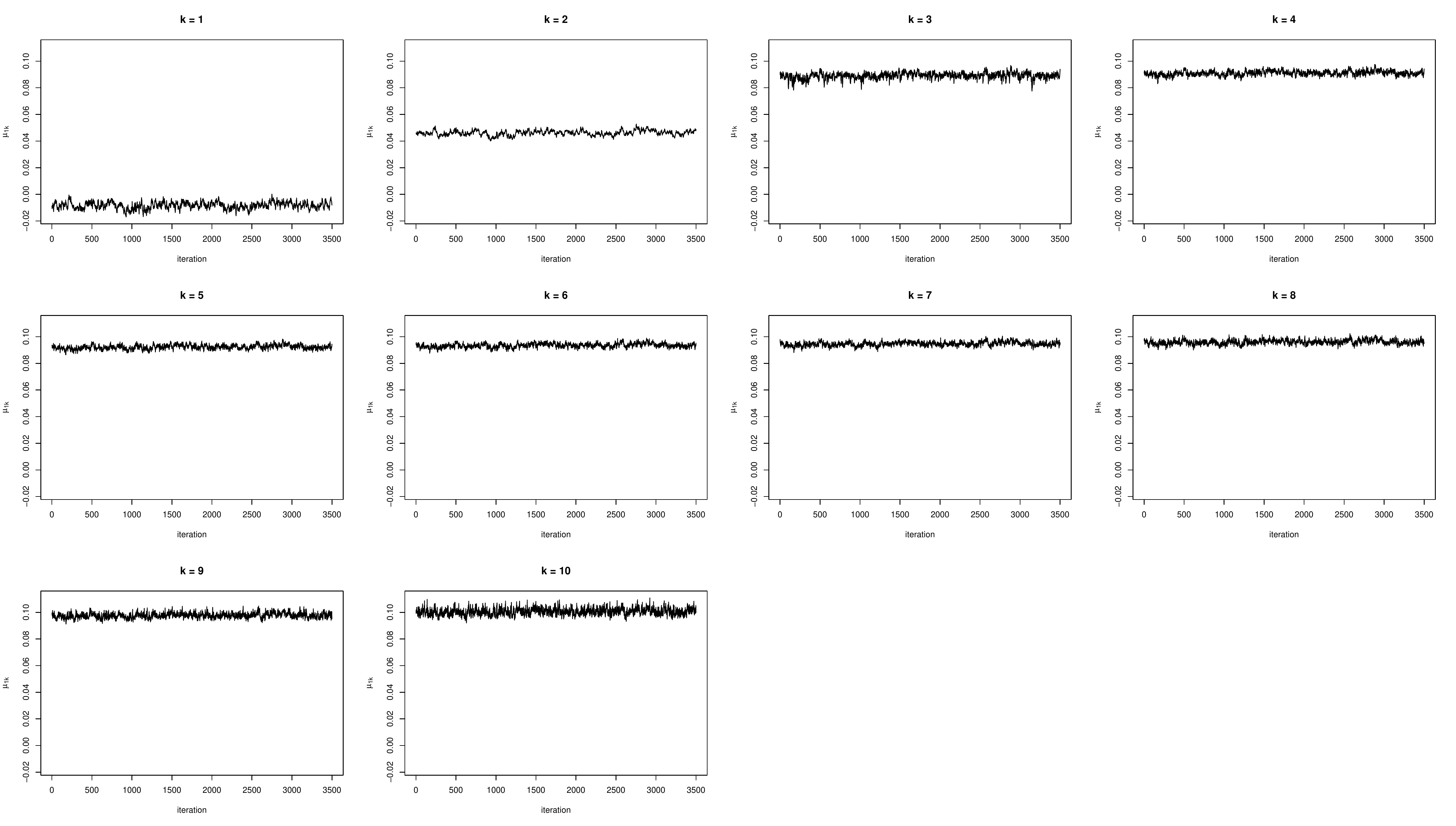}
\caption{Trace plots of $\mu_{1k}$. At each iteration, the means were sorted in increasing order to avoid label switching.}
\label{fig:mu1_traceplot}
\end{figure} 

\begin{figure}[h]
\centering
\includegraphics[scale=0.35]{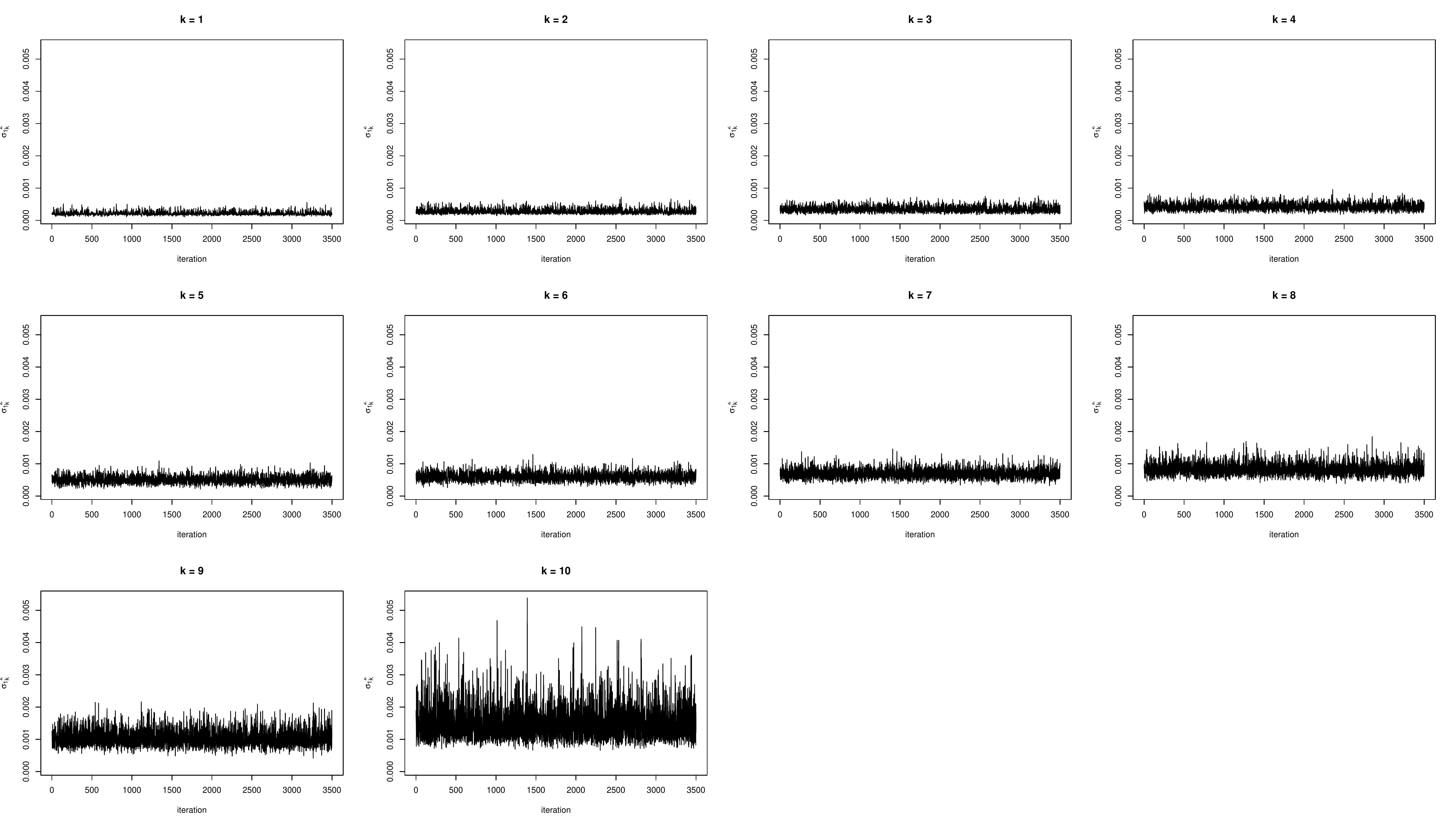}
\caption{Trace plots of $\sigma_{1k}^2$. At each iteration, the variances were sorted in increasing order to avoid label switching.}
\label{fig:sigma1_traceplot}
\end{figure} 

\begin{figure}[h]
\centering
\includegraphics[scale=0.35]{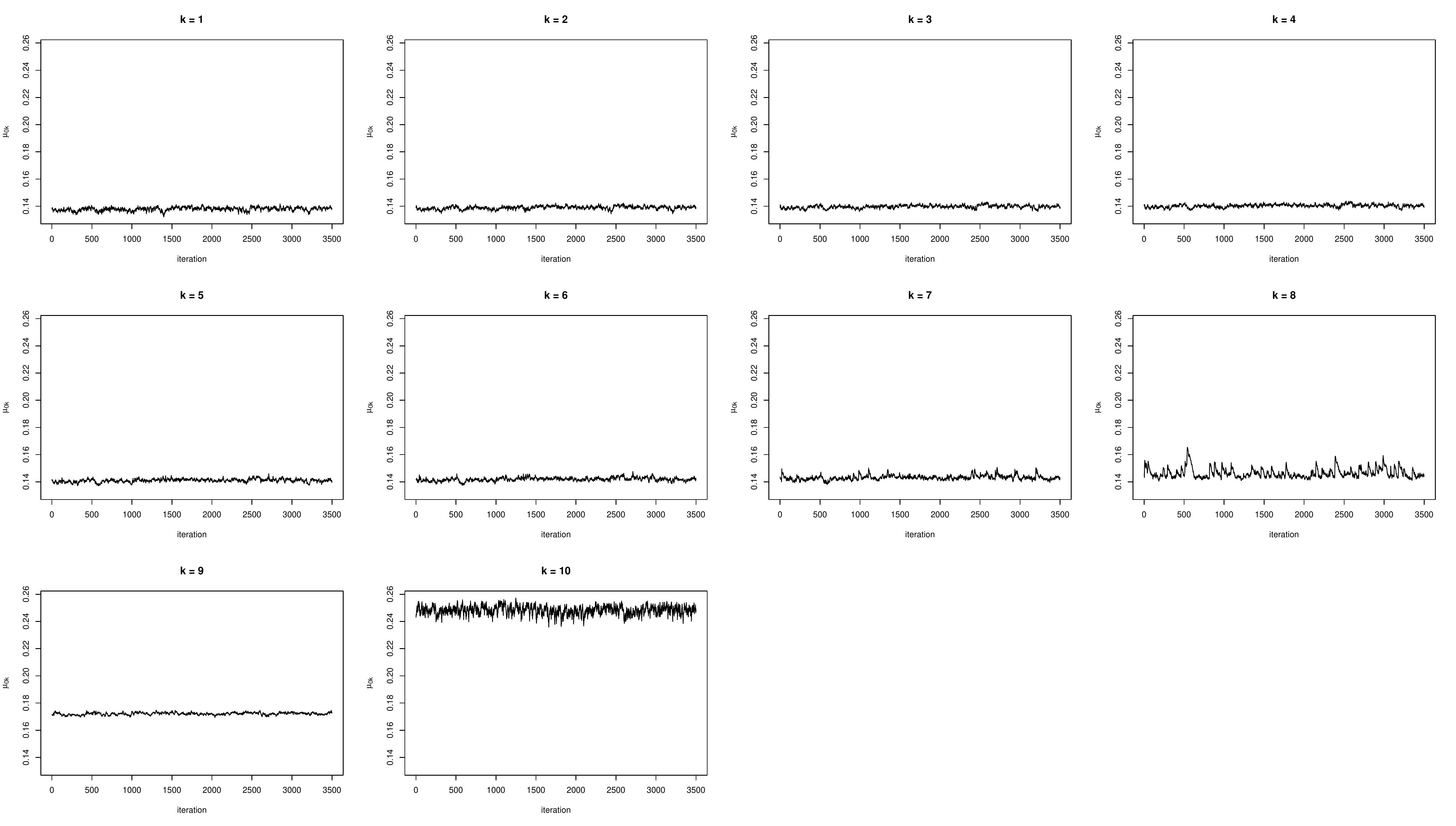}
\caption{Trace plots of $\mu_{0k}$. At each iteration, the means were sorted in increasing order to avoid label switching.}
\label{fig:mu0_traceplot}
\end{figure} 

\begin{figure}[h]
\centering
\includegraphics[scale=0.35]{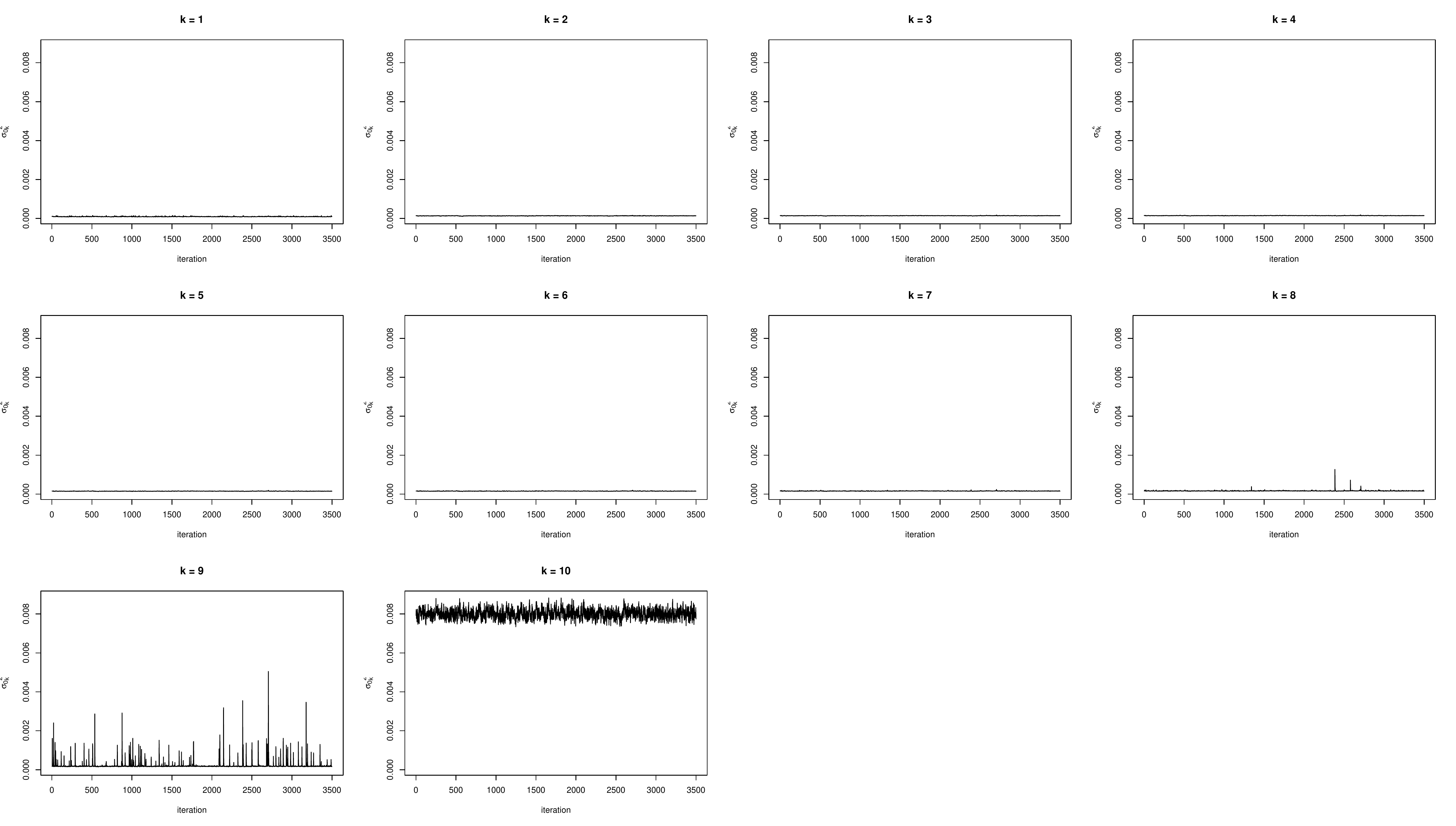}
\caption{Trace plots of $\sigma_{0k}^2$. At each iteration, the variances were sorted in increasing order to avoid label switching.}
\label{fig:sigma0_traceplot}
\end{figure} 

\begin{figure}[h]
\centering
\includegraphics[scale=0.35]{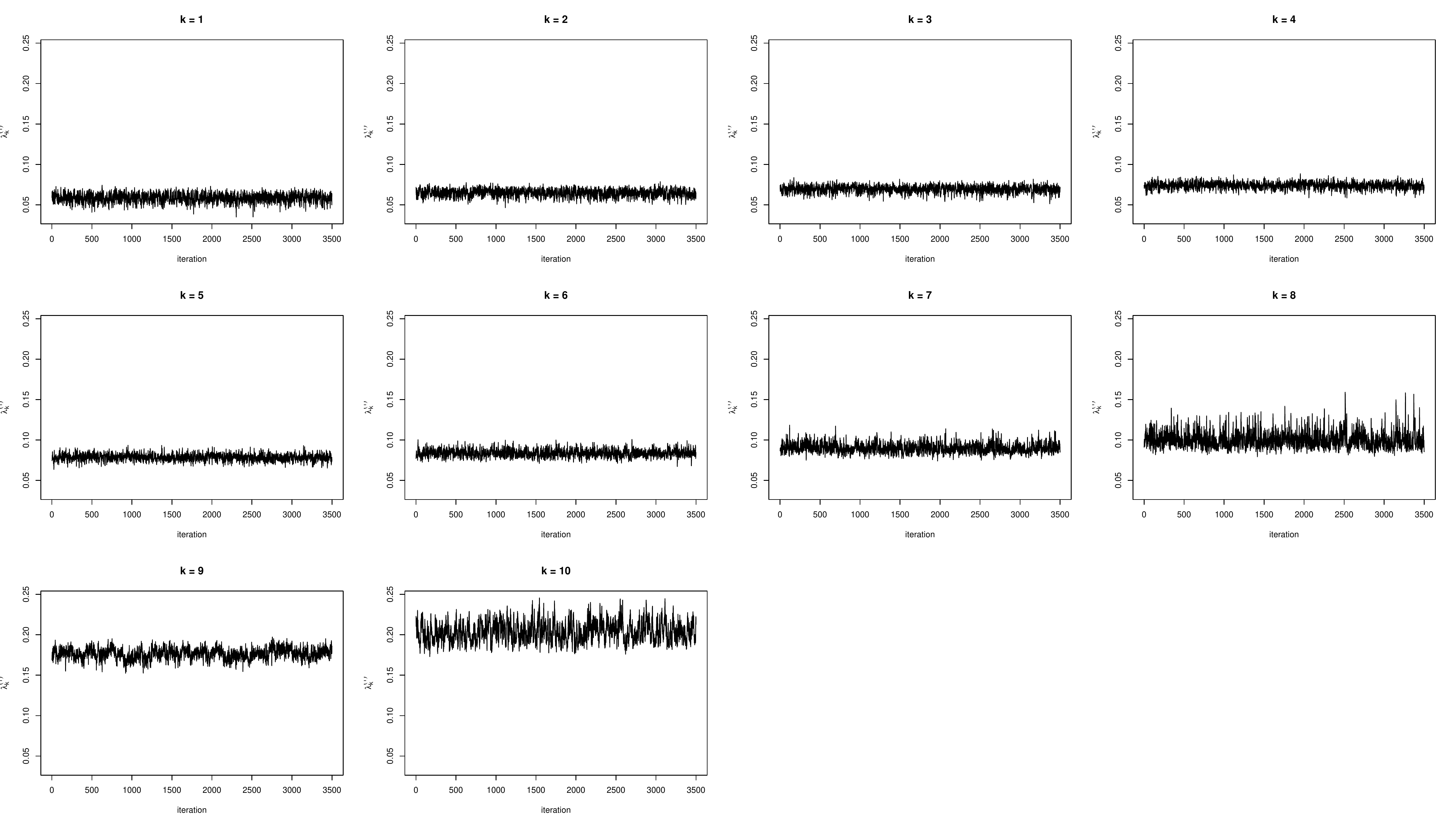}
\caption{Trace plots of $\lambda_{k}^{(1)}$.}
\label{fig:pk1_traceplot}
\end{figure} 

\begin{figure}[h]
\centering
\includegraphics[scale=0.35]{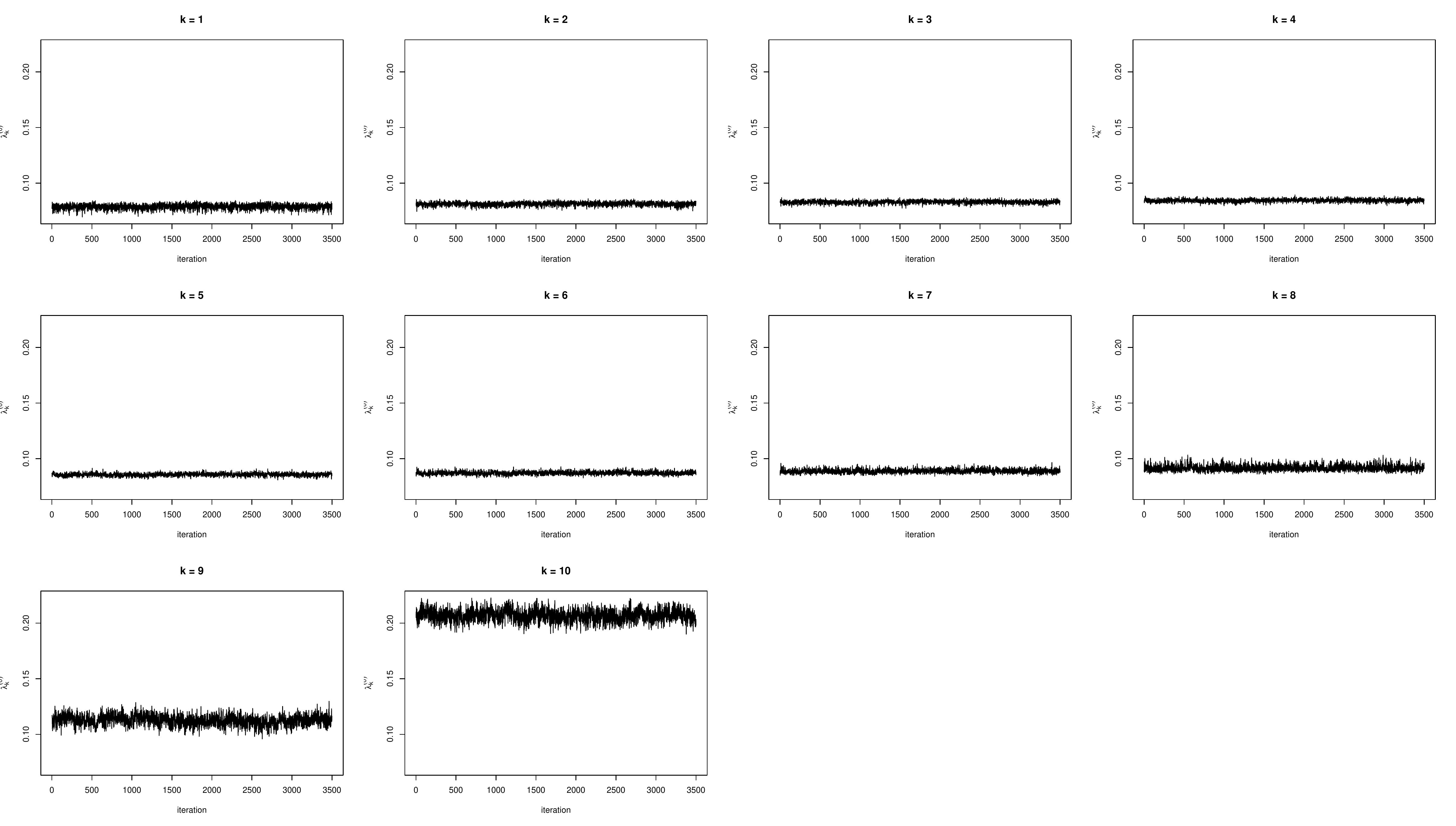}
\caption{Trace plots of $\lambda_{k}^{(0)}$.}
\label{fig:pk0_traceplot}
\end{figure}

\clearpage
\bibliographystyle{plain}
\bibliography{BHTSsupbibl}

%% file: model.pdf_t
\begin{picture}(0,0)%
\includegraphics{model.pdf}%
\end{picture}%
\setlength{\unitlength}{4144sp}%
\begingroup\makeatletter\ifx\SetFigFont\undefined%
\gdef\SetFigFont#1#2#3#4#5{%
  \reset@font\fontsize{#1}{#2pt}%
  \fontfamily{#3}\fontseries{#4}\fontshape{#5}%
  \selectfont}%
\fi\endgroup%
\begin{picture}(9564,5379)(-56,-6778)
\put(1891,-2311){\makebox(0,0)[lb]{\smash{{\SetFigFont{12}{14.4}{\rmdefault}{\mddefault}{\updefault}$\mu_{10}$}}}}
\put(7291,-2311){\makebox(0,0)[lb]{\smash{{\SetFigFont{12}{14.4}{\rmdefault}{\mddefault}{\updefault}$\mu_{00}$}}}}
\put(3286,-3256){\makebox(0,0)[lb]{\smash{{\SetFigFont{12}{14.4}{\rmdefault}{\mddefault}{\updefault}$\sigma_1^2$}}}}
\put(5986,-3256){\makebox(0,0)[lb]{\smash{{\SetFigFont{12}{14.4}{\rmdefault}{\mddefault}{\updefault}$\sigma_0^2$}}}}
\put(1936,-3256){\makebox(0,0)[lb]{\smash{{\SetFigFont{12}{14.4}{\rmdefault}{\mddefault}{\updefault}$\mu_1$}}}}
\put(7336,-3256){\makebox(0,0)[lb]{\smash{{\SetFigFont{12}{14.4}{\rmdefault}{\mddefault}{\updefault}$\mu_0$}}}}
\put(2566,-3931){\makebox(0,0)[lb]{\smash{{\SetFigFont{12}{14.4}{\rmdefault}{\mddefault}{\updefault}$\theta_{mi}^{(1)}$}}}}
\put(1711,-4831){\makebox(0,0)[lb]{\smash{{\SetFigFont{12}{14.4}{\rmdefault}{\mddefault}{\updefault}$\alpha_1$}}}}
\put(8371,-3931){\makebox(0,0)[lb]{\smash{{\SetFigFont{12}{14.4}{\rmdefault}{\mddefault}{\updefault}$G_0^{(0)}$}}}}
\put(721,-3931){\makebox(0,0)[lb]{\smash{{\SetFigFont{12}{14.4}{\rmdefault}{\mddefault}{\updefault}$G_0^{(1)}$}}}}
\put(811,-3031){\makebox(0,0)[lb]{\smash{{\SetFigFont{12}{14.4}{\rmdefault}{\mddefault}{\updefault}$\tau_1$}}}}
\put(7561,-4831){\makebox(0,0)[lb]{\smash{{\SetFigFont{12}{14.4}{\rmdefault}{\mddefault}{\updefault}$\alpha_0$}}}}
\put(8461,-3031){\makebox(0,0)[lb]{\smash{{\SetFigFont{12}{14.4}{\rmdefault}{\mddefault}{\updefault}$\tau_0$}}}}
\put(1621,-3931){\makebox(0,0)[lb]{\smash{{\SetFigFont{12}{14.4}{\rmdefault}{\mddefault}{\updefault}$G_{mi}^{(1)}$}}}}
\put(6616,-3931){\makebox(0,0)[lb]{\smash{{\SetFigFont{12}{14.4}{\rmdefault}{\mddefault}{\updefault}$\theta_{mi}^{(0)}$}}}}
\put(7471,-3931){\makebox(0,0)[lb]{\smash{{\SetFigFont{12}{14.4}{\rmdefault}{\mddefault}{\updefault}$G_{mi}^{(0)}$}}}}
\put(4591,-3931){\makebox(0,0)[lb]{\smash{{\SetFigFont{12}{14.4}{\rmdefault}{\mddefault}{\updefault}$z_{mi}$}}}}
\put(4591,-4831){\makebox(0,0)[lb]{\smash{{\SetFigFont{12}{14.4}{\rmdefault}{\mddefault}{\updefault}$b_{mi}$}}}}
\put(4681,-5731){\makebox(0,0)[lb]{\smash{{\SetFigFont{12}{14.4}{\rmdefault}{\mddefault}{\updefault}$\pi$}}}}
\put(4186,-6586){\makebox(0,0)[lb]{\smash{{\SetFigFont{12}{14.4}{\rmdefault}{\mddefault}{\updefault}$a_{\pi}$}}}}
\put(5131,-6586){\makebox(0,0)[lb]{\smash{{\SetFigFont{12}{14.4}{\rmdefault}{\mddefault}{\updefault}$b_{\pi}$}}}}
\put(361,-2086){\makebox(0,0)[lb]{\smash{{\SetFigFont{12}{14.4}{\rmdefault}{\mddefault}{\updefault}$a_{\tau}$}}}}
\put(1306,-2086){\makebox(0,0)[lb]{\smash{{\SetFigFont{12}{14.4}{\rmdefault}{\mddefault}{\updefault}$b_{\tau}$}}}}
\put(8011,-2086){\makebox(0,0)[lb]{\smash{{\SetFigFont{12}{14.4}{\rmdefault}{\mddefault}{\updefault}$a_{\tau}$}}}}
\put(8956,-2086){\makebox(0,0)[lb]{\smash{{\SetFigFont{12}{14.4}{\rmdefault}{\mddefault}{\updefault}$b_{\tau}$}}}}
\put(3781,-2356){\makebox(0,0)[lb]{\smash{{\SetFigFont{12}{14.4}{\rmdefault}{\mddefault}{\updefault}$b$}}}}
\put(2881,-2356){\makebox(0,0)[lb]{\smash{{\SetFigFont{12}{14.4}{\rmdefault}{\mddefault}{\updefault}$a$}}}}
\put(6436,-2356){\makebox(0,0)[lb]{\smash{{\SetFigFont{12}{14.4}{\rmdefault}{\mddefault}{\updefault}$b$}}}}
\put(5581,-2356){\makebox(0,0)[lb]{\smash{{\SetFigFont{12}{14.4}{\rmdefault}{\mddefault}{\updefault}$a$}}}}
\put(1216,-5686){\makebox(0,0)[lb]{\smash{{\SetFigFont{12}{14.4}{\rmdefault}{\mddefault}{\updefault}$a_{\alpha}$}}}}
\put(2116,-5686){\makebox(0,0)[lb]{\smash{{\SetFigFont{12}{14.4}{\rmdefault}{\mddefault}{\updefault}$b_{\alpha}$}}}}
\put(8011,-5686){\makebox(0,0)[lb]{\smash{{\SetFigFont{12}{14.4}{\rmdefault}{\mddefault}{\updefault}$b_{\alpha}$}}}}
\put(7111,-5686){\makebox(0,0)[lb]{\smash{{\SetFigFont{12}{14.4}{\rmdefault}{\mddefault}{\updefault}$a_{\alpha}$}}}}
\end{picture}%